\documentclass{article} % For LaTeX2e
\usepackage{arxiv}
\usepackage{url}            % simple URL typesetting
\usepackage{booktabs}       % professional-quality tables
\usepackage{multirow}    
\usepackage{amsfonts}       % blackboard math symbols
\usepackage{nicefrac}       % compact symbols for 1/2, etc.
\usepackage{microtype}      % microtypogrhy
\usepackage{natbib}
\usepackage{enumerate}
\usepackage{hhline}
\usepackage{makecell}
\usepackage{pifont}
\usepackage{xspace}

\usepackage{graphicx} % more modern
\usepackage{caption}
\usepackage{subcaption}
\usepackage{amsmath}
\usepackage{amsthm}
\usepackage{amssymb}
\usepackage{tikz}
\usepackage{xcolor}
\usetikzlibrary{arrows}

\allowdisplaybreaks

\usepackage{mathrsfs}

\usepackage{hyperref}
\usepackage{bm}
\allowdisplaybreaks

\newcommand{\lnorm}{\left\Vert}
\newcommand{\rnorm}{\right\Vert}

\newcommand{\expect}{\mathbb{E}}

\newcommand{\indict}{\mathbb{I}}

\newtheorem{lem}{Lemma}
\newtheorem{prop}{Proposition}

\usepackage[capitalize,noabbrev]{cleveref}
\crefname{thm}{Theorem}{Theorems}
\crefname{lem}{Lemma}{Lemmas}
\crefname{cor}{Corollary}{Corollaries}
\crefname{prop}{Proposition}{Propositions}
\crefname{asmp}{Assumption}{Assumptions}
\crefname{defn}{Definition}{Definitions}
\crefname{oracle}{Oracle}{Oracles}
\crefname{fact}{Fact}{Facts}
\crefname{conj}{Conjecture}{Conjectures}
\crefname{rem}{Remark}{Remarks}
\crefname{example}{Example}{Examples}
\crefname{condition}{Condition}{Conditions}
\crefname{exercise}{Exercise}{Exercises}
\crefname{table}{Table}{Tables}
\crefname{figure}{Figure}{Figures}
\crefname{section}{Section}{Sections}
\crefname{subsection}{Section}{Sections}
\crefname{appendix}{Appendix}{Appendices}
\crefname{message}{Message}{Messages}

\definecolor{red}{rgb}{1, 0, 0}

\definecolor{green}{rgb}{0, 1, 0}

\definecolor{blue}{rgb}{0, 0, 1}

\definecolor{orange}{rgb}{1, 0.4, 0.0}

\newcommand{\reals}{{\mathbb{R}}}

\newcommand{\argmin}{\mathop{\rm argmin}}

\usepackage{mathtools}
\usepackage{amsmath}
\usepackage{amsthm}
\usepackage{amsfonts}
\usepackage{caption}
\usepackage{subcaption}
\usepackage{xspace}
\usepackage{color}
\usepackage{dcolumn}
\usepackage{makecell}
\newcommand{\secspace}{\vspace{-0.1cm}}

\usepackage{wrapfig}
\usepackage{multirow}
\usepackage{natbib}
\definecolor{mine}{RGB}{205, 232, 248}%
\usepackage{caption}
\renewcommand{\eqref}[1]{Eq.(\ref{#1})}
\crefformat{figure}{Fig.~#2#1#3}
\Crefformat{figure}{Figure~#2#1#3}
\usepackage{hyperref}
\usepackage{graphicx}
\input{packages/math_commands.tex}
\usepackage{url}
\newcommand{\ours}{MOREC\xspace}
\usepackage[ruled, vlined, linesnumbered]{algorithm2e}

\title{Reward-Consistent Dynamics Models are Strongly Generalizable for Offline Reinforcement Learning}

\usepackage{authblk}
\newcommand\nnfootnote[1]{%
  \begin{NoHyper}
  \renewcommand\thefootnote{}\footnote{#1}%
  \addtocounter{footnote}{-1}%
  \end{NoHyper}
}
\author[ ]{\textbf{Fan-Ming Luo}}
\author[ ]{\textbf{Tian Xu}}
\author[ ]{\textbf{Xingchen Cao}}
\author[ ]{\textbf{Yang Yu}\textsuperscript{\dag}}
\affil[ ]{National Key Laboratory for Novel Software Technology, Nanjing University, China}
\affil[ ]{Polixir.ai}
\affil[ ]{\texttt{\{luofm,xut,caoxc,yuy\}@lamda.nju.edu.cn}}
\date{}
\renewcommand{\headeright}{}
\renewcommand{\shorttitle}{}

\begin{document}

\maketitle

\begin{abstract}
Learning a precise dynamics model can be crucial for offline reinforcement learning, which, unfortunately, has been found to be quite challenging. Dynamics models that are learned by fitting historical transitions often struggle to generalize to unseen transitions. In this study, we identify a hidden but pivotal factor termed \emph{dynamics reward} that remains consistent across transitions, offering a pathway to better generalization. Therefore, we propose the idea of reward-consistent dynamics models: any trajectory generated by the dynamics model should maximize the dynamics reward derived from the data. We implement this idea as the MOREC (Model-based Offline reinforcement learning with Reward Consistency) method, which can be seamlessly integrated into previous offline model-based reinforcement learning (MBRL) methods. MOREC learns a generalizable dynamics reward function from offline data, which is subsequently employed as a transition filter in any offline MBRL method: when generating transitions, the dynamics model generates a batch of transitions and selects the one with the highest dynamics reward value. On a synthetic task, we visualize that MOREC has a strong generalization ability and can surprisingly recover some distant unseen transitions. On 21 offline tasks in D4RL and NeoRL benchmarks, MOREC improves the previous state-of-the-art performance by a significant margin, i.e., 4.6\% on D4RL tasks and 25.9\% on NeoRL tasks. Notably, MOREC is the first method that can achieve above 95\% online RL performance in 6 out of 12 D4RL tasks and 3 out of 9 NeoRL tasks.

\end{abstract}

\nnfootnote{\dag: Yang Yu is the corresponding author.}

\section{Introduction}

Model-based approaches in reinforcement learning (MBRL) encompass techniques that harness either an established environment model or one learned to approximate the environment, thereby addressing RL challenges~\citep{sutton1998rl,luo2022survey, moerland2023mbrlsurvey}. These models are primarily employed to forecast forthcoming states arising from the execution of actions within the environment. By leveraging models, it becomes possible to evaluate action sequences or policies through simulation, circumventing the need for direct interactions with the actual environment and substantially curtailing sampling expenses. As a result, model-based methods empower offline RL scenarios where agents exclusively operate with a dataset sampled from the environment, devoid of direct access to the environment itself, thus enabling the adoption of a range of efficient methodologies~\citep{yu2020mopo,yu2021combo, rigter2022rambo, sum2023mobile}.

One can conduct numerous trial searches within an ideal model until the optimal policies for a task are discovered. However, it is incredibly challenging to learn a uniformly accurate dynamics model solely through initial supervised learning on the offline dataset. Without additional guidance on models, model errors will inevitably occur. These errors become particularly pronounced in state-action pairs that fall outside the distribution of the offline data. Such instances may arise when dealing with long forecasting horizons or when the learning policies deviate significantly from the behavior policy of the offline data. These model errors have the potential to be erroneously exploited by algorithms, consequently yielding policies that exhibit significant underperformance.

To deal with the model error, there are mainly two branches of research, i.e., developing policy learning approaches that bypass the model error and developing better model learning approaches.

MOPO~\citep{yu2020mopo} was the first to propose penalizing rewards based on estimated model uncertainties and limiting the rollout horizon during model simulations. This strategy reduces the deviation of learning policies from the behavior policy, thereby minimizing visits to out-of-distribution (OOD) state-action pairs. Subsequent works in this direction propose more advanced conservative techniques. MORel~\citep{lar2020morel} introduced an early stopping mechanism to avoid model exploitation. RAMBO~\citep{rigter2022rambo} developed pessimistic models as an alternative to uncertainty estimation. MOBILE~\citep{sum2023mobile} improved model uncertainty estimation to broaden the scope of policy exploration. It is important to note that there are also non-conservative approaches. MAPLE~\citep{chen2021maple} introduced contexture meta-policy learning in models to enable generalization to unseen situations. However, despite these efforts, the performance of these techniques is inherently limited by the capability of the learned model itself.

To develop better model learning approaches, VirtualTaobao~\citep{shi2019taobao} first proposed adversarial model learning to achieve better models, which was also applied in more complex industrial applications~\citep{shang2021mlj}. Adversarial model learning was then shown to address the compounding error issue~\citep{xu2022model_imitation,xu2022understanding,xu2023provably}, allowing for the use of models in rolling out long trajectories. Adversarial counterfactual model learning~\citep{chen2022galileo} demonstrated the ability to learn causal transitions that are difficult to capture through simple supervised learning.

This paper firstly focuses on learning better models. To enhance the generalization ability of dynamics models, our goal is to identify the invariance underlying the data. We identify a hidden but pivotal factor named \textbf{dynamics reward}, which represents the inherent driving forces of the environment dynamics. For instance, a naive dynamics reward is a function that assigns 1 to true transitions and 0 to false transitions. We can conceptualize the dynamics model as an \emph{agent} which maximizes the dynamics reward~\citep{xu2022model_imitation}. We then propose the idea of reward-consistent dynamics models: any trajectory generated by the dynamics model should maximize the dynamics reward derived from the data. To implement this idea, we learn a dynamics reward function through an inverse reinforcement learning (IRL) method from the offline data. We finally incorporate the dynamics reward into the policy learning stage by modifying the model-rollout procedure. Specifically, we (i) regulate the model's output to generate rollouts associated with higher dynamics rewards; (ii) terminate the rollout upon encountering states with low dynamics rewards. This method can be integrated into most prior offline MBRL methods, resulting in a novel one named \ours (Model-based Offline policy optimization with REward Consistency).

In experiments, we first evaluate \ours on a synthetic controlling task. We visualize that the learned dynamics reward aligns closely with the accuracy of transitions, even in OOD regions, demonstrating its superior generalization ability. Furthermore, we demonstrate that the utilization of such a generalizable dynamics reward facilitates the generation of high-fidelity model rollouts, leading to a substantial enhancement in policy performance. Additionally, we evaluate \ours on 21 typical tasks from two offline benchmarks, D4RL~\citep{fu2020d4rl} and NeoRL~\citep{qin2022neorl}. The empirical results show \ours outperforms prior state-of-the-art (SOTA) methods on 18 out of the 21 tasks. Notably, in the more difficult NeoRL benchmark, \ours achieves a remarkable $25.9\%$ average performance improvement over prior SOTAs and solves $3$ tasks for the first time ($0$ previously). In-depth analysis shows a strong correlation between the learned dynamics rewards and model prediction errors. Guided by such an instructive dynamics reward, \ours shows a significant reduction of the model prediction errors even when a long rollout horizon is adopted, enabling us to increase the rollout horizon to $100$ in all experiments.

\section{Preliminaries and Related Work}
% \subsection{Offline Model-Based Reinforcement Learning}
\noindent{\textbf{Reinforcement Learning (RL).}} In RL, we consider the Markov decision process (MDP)~\citep{sutton1998rl}, described by a tuple $\langle\mathcal{S}, \mathcal{A}, P^\star, r^{\text{task}}, \gamma, \rho_0\rangle$, 
where $\mathcal{S}$ is the state space, 
$\mathcal{A}$ is the action space, $P^\star: \gS \times \gA \rightarrow \Delta (\gS)$ is the true dynamics, 
$r^{\text{task}}: \mathcal{S}\times\mathcal{A}\rightarrow \mathbb{R}$ is the task reward function,
$\gamma\in(0,1)$ is the discount factor, 
and
$\rho_0$ is the initial state distribution.

The value function $V^\pi(s) = \mathbb{E}_\pi\left[ \sum_{i=0}^\infty\gamma^i r^{\text{task}} (s_{t+i},a_{t+i})\mid s_t = s\right]$ is the expected cumulative rewards when starting at state $s_t=s$ and following $\pi$. 
Similarly, we can define the action-value function as $Q^\pi(s,a) = \mathbb{E}_\pi\left[ \sum_{i=0}^\infty\gamma^i r^{\text{task}} (s_{t+i},a_{t+i}) \mid s_t = s, a_t = a\right]$. 
The objective of RL is to find a policy $\pi$ that maximizes the expected value under the initial state distribution, i.e., $\mathop{\max}_{\pi} ~ \mathbb{E}_{ s_0 \sim \rho_0(\cdot)} \left[V^\pi(s_0)\right]$.

\noindent{\textbf{Offline RL.}}
Offline RL~\citep{levine2020offline} emphasizes learning from an offline dataset without additional interaction with the environment. Here the offline dataset $\gD = \{ (s, a, r^{\text{task}}, s^\prime) \}$ consists of transitions from trajectories collected by a behavior policy. A primary obstacle faced in this domain arises from discrepancies between the offline data and the behavior policy, which leads to extrapolation errors~\citep{kumar2019stabilizing}. To mitigate these errors, model-free offline RL methods have integrated conservatism, either by modulating the policy or the Q-function of online RL algorithms~\citep{fujimoto2019off, bai2022pessimistic}. 

\noindent{\textbf{Offline MBRL.}}
Building upon the foundational concepts of offline RL, offline MBRL introduces an advanced strategy by leveraging a dynamics model constructed from offline data. 
Offline MBRL methods typically possess two stages: (i) learn a model of the environment from the offline data $\mathcal{D}$; (ii) learn a policy from the model and $\mathcal{D}$. Let the model parameterized by $\theta$ be $P_\theta(s_{t+1} | s_t,a_t)$. It will be trained by log-likelihood maximization:
\begin{equation}
\label{eq_model_loss}
\mathop{\max}_{\theta}~~\gL_M (\theta) :=\mathbb{E}_{ (s, a, s^\prime) \sim \mathcal{D}}\left[\log( P_{\theta} (s^\prime | s, a)) \right].
\end{equation}
Without loss of generality, we assume that the task reward function $r^{\text{task}}$ is known because $r^{\text{task}}$ can be considered as part of the dynamics model. The strategy of offline MBRL allows for the generation of synthetic data, potentially enhancing adaptability to out-of-distribution states~\citep{yu2020mopo, lar2020morel}. However, the learned dynamics model also inevitably suffers from errors for the limited experience dataset~\citep{xu2022model_imitation, michael2019mbpo}. To mitigate the challenges stemming from model inaccuracies, several approaches integrate conservatism through methods like uncertainty estimation to guide the behavior policy closer to the available dataset~\citep{yu2020mopo}. The most recent works focus on designing better conservative strategies~\citep{lar2020morel,sum2023mobile,rigter2022rambo} to unleash the full potential of the model. However, these methods are still limited by the static capabilities of the learned model itself.
In this paper, we present a novel approach to enhance the fidelities of model rollouts, through guidence of reward signals. This method can not only seamlessly integrate seamlessly but also augment previous offline MBRL techniques.

\noindent{\textbf{Inverse Reinforcement Learning (IRL).}}
IRL \citep{ng2000irl, ni2020firl} aims to recover a reward function from expert demonstrations. A main class of algorithms \citep{ng2004apprentice, ho2016gail, swamy2021moments, luo2022darl}  in IRL infer a reward function by maximizing the value gaps between the expert policy and the other policies, as the expert policy should perform well under the desired reward function. It is worth noting that previous IRL methods recover a reward function for a policy. In this paper, we apply IRL to learn a reward function for the dynamics model.

\secspace
\begin{figure*}[t]
\centering 
\includegraphics[width=1.0 \linewidth]{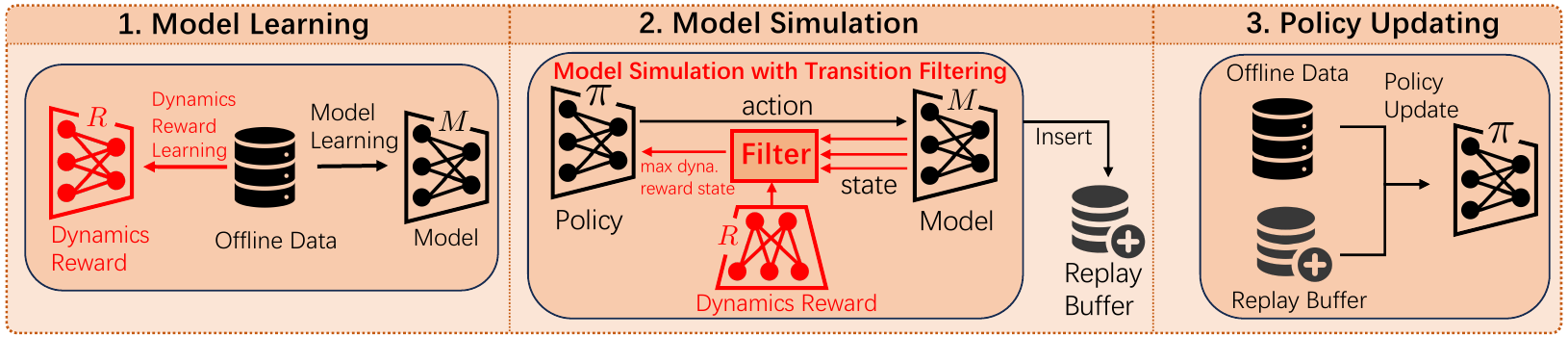}
\caption{Framework of \ours. Key distinctions between \ours and the prior offline MBRL are emphasized in red. The main modifications arise in the model learning and model simulation stages.} 
\label{fig_framework}  
\end{figure*}

\section{Method}
\label{sec_method}
\secspace
In this section, we will delve into the details of how we learn the dynamics reward function and how we leverage this reward function to facilitate generating high-fidelity model rollouts. We begin with an overview of our method in Section~\ref{sec_framework}, followed by an explanation of dynamics reward learning in Section~\ref{sec_reward_learning}. At last, we present the complete process of \ours in Section~\ref{sec_morec}.

\subsection{Overview of \ours}
\label{sec_framework}

We aim to improve the fidelity of model rollouts along with the policy improvement. 
To accomplish this, we must uncover the underlying invariance across different data instances. We have identified a crucial hidden factor called dynamics reward, which represents the intrinsic driving forces of the environment dynamics. By considering the dynamics system as an agent~\citep{xu2022model_imitation}, we can assume that the dynamics system was also learned by maximizing the dynamics reward. Therefore, regardless of the policy used to generate interaction data, the data should exhibit consistent dynamics reward.

%There are mainly two types of learning signals in machine learning: the label signal in supervised learning and the reward signal in RL. In offline RL, it is impossible to acquire additional environment transitions as the label signal. Therefore, we seek to use the reward signal to generate accurate model rollouts. 

Based on this idea, we develop a new offline MBRL framework \ours, which is illustrated in \cref{fig_framework}. As a general framework, \ours can be applied to the most existing model-based offline RL methods such as MOPO \citep{yu2020mopo}, MOREL \citep{lar2020morel}, and MOBILE \citep{sum2023mobile}. Compared with the framework of prior model-based offline RL methods, \ours incorporates two significant algorithmic designs, which are marked by red color in \cref{fig_framework}. First, we apply IRL to infer a dynamics reward function tailored for the dynamics model. We elaborate on the training process of the dynamics reward in Section \ref{sec_reward_learning}. The second algorithmic design is located in the model simulation stage. In particular, we utilize the learned dynamics reward to help generate high-fidelity model rollouts. The detailed generation procedure is explained in Section \ref{sec_morec}.

\subsection{Dynamics Reward Learning}
\label{sec_reward_learning}
In this part, we present how we learn a reward function specially designed for dynamics models. Our method is motivated by the new perspective of dynamics models. That is, we can view the dynamics model as a \emph{agent}~\citep{xu2022model_imitation} with the new augmented state space $\gS \times \gA$ and the new action space $\gS$. Accordingly, the behavior policy $\pi_{\beta}$ is regarded as the ``dynamics model''. Besides, there is a dynamics reward function $r^{D}: \gS \times \gA \times \gS \rightarrow \reals$ that can judge the performance of a dynamics model. For instance, a naive dynamics reward is a function that assigns $1$ on true transitions and $0$ on false transitions. Under the dynamics reward, the true dynamics is the \emph{policy} that achieves the maximum dynamics reward. From this perspective, we can formulate the problem of dynamics reward learning as an IRL problem, which aims to infer a reward function from expert demonstrations collected by the expert policy. In particular, the true dynamics is viewed as the expert policy and the offline dataset collected in the true dynamics is viewed as the expert demonstrations in IRL. Our target is to learn a reward function that can explain the true dynamics (i.e., the expert policy).

Based on the above IRL formulation, we leverage the advancements in IRL to help infer the dynamics reward. In particular, inspired by the well-known IRL-equivalent method GAIL \citep{ho2016gail}, we consider the following dynamics reward learning objective.
\begin{equation}
\label{eq_original_discriminator_loss}
    \ell (D, P) = \expect_{(s, a, s^\prime) \sim \gD} \ls \log (D (s, a, s^\prime)) \rs + \expect_{(s, a) \sim \gD, s^\prime \sim P (\cdot|s, a)} \ls \log (1-D(s, a, s^\prime)) \rs.
\end{equation}
With a slight abuse of notation, we use $\gD$ to denote both the state-action and state-action-next-state distributions from the offline dataset. Here $P: \gS \times \gA \rightarrow \Delta (\gS)$ is an arbitrary dynamics model and $D: \gS \times \gA \times \gS \rightarrow (0, 1)$ is a discriminator, which is used to induce the dynamics reward. More concretely, when we consider the dynamics reward of $r^{D} (s, a, s^\prime) = - \log (1 - D (s, a, s^\prime))$ \citep{ho2016gail}, $\ell (D, P)$ can be roughly interpreted as the difference between rewards on the true dynamics $P^\star$ and on the dynamics model $P$. Notice that a desired dynamics reward should induce a large reward difference $\ell (D, P)$ since the true dynamics achieves the maximal reward on such a dynamics reward function.    

Following the maximum margin IRL method \citep{ng2004apprentice, ratliff2006maximum, ho2016gail}, we further define the margin function $f (D) := \min_{P} \ell (D, P)$ and consider maximizing such a margin function.
\begin{align*}
    \max_{D} f (D) = \max_{D} \min_{P} \expect_{(s, a, s^\prime) \sim \gD} \ls \log (D (s, a, s^\prime)) \rs + \expect_{(s, a) \sim \gD, s^\prime \sim P (\cdot|s, a)} \ls \log (1-D(s, a, s^\prime)) \rs. 
\end{align*}
In other words, we aim to find a dynamics reward that can maximize the reward difference with an \textit{adversarial dynamics model} $P$. To optimize this objective, we apply the gradient-descent-ascent method \citep{lin2020gradient} which alternates between updating the dynamics reward and adversarial dynamics model. For the inner problem, we utilize the policy gradient method \citep{sutton1999policy} to update the adversarial dynamics model. The outer problem is exactly the binary classification problem by considering the transitions in $\gD$ as positive samples and the transitions generated by $P$ as negative samples. The detailed procedure is outlined in Algorithm \ref{alg_model_reward_learning} in Appendix \ref{appendix:model_reward_learning_algorithm}.

Note that, in the original field of imitation learning, the above IRL method would learn reward models that are coupled with dynamics~\citep{fu2017airl,luo2022darl} and thus may not generalize well. Following \cite{luo2022darl}, we employ an ensemble of discriminators as the dynamics reward, i.e., $\frac{1}{T}\sum_{t=1}^T D_t$, which can be theoretically justified by providing a convergence analysis.

\begin{prop}
\label{prop:convergence}
    Consider Algorithm \ref{alg_model_reward_learning}. Assume that the gradient norm is bounded, i.e., $\forall t \in [T], \lnorm \nabla_{P} \ell (D_t, P_t) \rnorm_2 \leq G^{P}, \; \lnorm \nabla_{D} \ell (D_t, P_t) \rnorm_2 \leq G^{D}$. If we take the step size of $\{\eta^{P}_t = \sqrt{|\gS| |\gA|} / (G^Pt) , \; \eta^D_t = \sqrt{|\gS|^2 |\gA|} / (G^D t)  \}_{t=1}^T$, then we have
 \begin{align*}
     f \lp \frac{1}{T} \sum_{t=1}^T D_t \rp \geq \max_{D} f (D) - \frac{3}{2} G^D \sqrt{\frac{|\gS|^2 |\gA|}{T}} - \frac{3}{2} G^{P} \sqrt{\frac{|\gS| |\gA|}{T}}.
 \end{align*}
\end{prop}
Please refer to Appendix~\ref{app_proposition_prove} for a thorough proof. \cref{prop:convergence} indicates that the proposed dynamics reward learning algorithm can converge to the global optimum with a rate of $\gO (1/\sqrt{T})$. Notice that \cref{prop:convergence} shows an average-iterate convergence \citep{nesterov2003introductory}. To consist with the theory, in practice, we maintain an ensemble of discriminators in the training procedure and output the dynamics reward of
\begin{align}
\label{eq_reward_representation}
    r^{D} (s, a, s^\prime) = - \log \lp 1 - \frac{1}{T} \sum_{t=1}^T D_t (s, a, s^\prime) \rp, \; \forall (s, a ,s^\prime) \in \gS \times \gA \times \gS.
\end{align}

\subsection{Model-Based Offline Policy Optimization with Reward Consisteny}
\label{sec_morec}
\begin{algorithm}[t]
\caption{\ours-MOPO} \label{alg_policy_updating} %\small
\KwIn{Offline data $\mathcal{D}$; initialized policy $\pi_{\phi}$ and critic $Q_{\psi}$; uncertainty penalty coefficient $\beta$;
%\RED{policy learning maximum rollout horizon $T$;}
}
\colorbox{mine}{Invoke Algorithm \ref{alg_model_reward_learning} to learn the dynamics reward $r^{D}$;}\\
Learn a model ensemble $\mathcal{P}=\{p_{\theta_i} (s'| s,a), \forall i\in[M]\}$ via \eqref{eq_model_loss} \;
Initialize an empty buffer $\mathcal{B} \leftarrow \phi$\;
\For {$i =  1, \dots, N_{\text{iter}}$}
{
\For{ $j=1,\dots, N_{\text{rollout}}$ }{
Sample a state $s_t$ from $\mathcal{D}$ uniformly \;
\colorbox{mine}{Generate a rollout $l=\{s_t,a_t,r^{\text{raw}}_t,...\}$ following \eqref{eq_ours_tc_generating} with the dynamics reward $r^D$;} \\
Minus the rewards with uncertainty penalties $r_t \leftarrow r^{\text{raw}}_t - \beta \mathcal{U}(\mathcal{P}, s_t,a_t,s_{t+1}), \forall r^{\text{raw}}_t\in l$\;\label{mopo_line_uncertainty}
Insert $\{s_t,a_t,r_t,s_{t+1},\dots\}$ to $\mathcal{B}$\;
}
Draw samples from $\gD \cup \gB$ and apply SAC to update $\pi_\phi$ and $Q_{\psi}$\;
}
\end{algorithm}
In this part, we elaborate on how MOREC incorporates the dynamics reward for guiding the dynamics model in generating model rollouts. The initial learning of dynamics models in MOREC is the same as that in the prior offline MBRL framework. In particular, we learn a model ensemble, denoted as $\{ P_{\theta_m} (s'| s, a), \; \forall m \in[M] \}$, where $\theta_m$ represents the parameter of the $m$-th model, $M$ symbolizes the ensemble size and $[M]:= \{1, 2, \cdots, M \}$. Each model in this ensemble is learned through supervised learning using \eqref{eq_model_loss}.

When using the learned dynamics models to generate samples, MOREC incorporates the dynamics reward to select high-fidelity transitions from all sampled candidates, which is the key difference from other existing model-based offline RL methods. More specifically, at time step $t$, \ours begins by sampling from each transition distribution $P_{\theta_i} (\cdot, | s_t,a_t)$ $N$ times to obtain $MN$ candidate of next states. The selections are then made based on a probability distribution induced by the softmax function, which favors transitions with higher dynamics rewards. The detailed rollout generation process is shown in \eqref{eq_ours_tc_generating}. 
\begin{equation}
    \label{eq_ours_tc_generating}
    \begin{aligned}
    &a_t\sim\pi_{\phi} (\cdot | s_t);~~s_{t+1}^{m,n} \sim P_{\theta_m} (\cdot | s_t,a_t), \; \forall m\in[M],n\in[N];~~s_{t+1} = s_{t+1}^{m^\star,n^\star};\\
    % &\\
    \text{where}~& m^\star,n^\star \sim P (\cdot), \; P (m, n) =  \frac{\exp(r^D(s_t,a_t,s_{t+1}^{m,n})/\kappa)}{\sum_{m'\in[M]}\sum_{n'\in[N]}\exp(r^D(s_t,a_t,s_{t+1}^{m',n'})/\kappa)}.\\
    % &.
    % \; r^\text{raw}_{t} = r_{t}^{m,n},\\
    \end{aligned}
\end{equation}
Here $\pi_{\phi}$ is the current learning policy with parameter $\phi$ and $\kappa$ is the temperature coefficient of the softmax function. The above process is repeated for $H_{\text{rollout}}$ times to generate a trajectory with the maximal length $H_{\text{rollout}}$. Besides,
in order to manage cases where the chosen transitions deviate significantly from the true one, leading to an extremely low dynamics reward, early rollout termination is also implemented when $r^D(s_t,a_t,s_{t+1})<r^D_{\min}$, where $r^D_{\min}$ is a preset hyperparameter. With the above transition filtering technique, we can leverage the dynamics reward to generate high-fidelity transitions, which benefit the subsequent policy learning phase.

As for the policy learning phase, the developed framework MOREC can incorporate the policy learning techniques in the most existing offline MBRL methods. In Algorithm \ref{alg_policy_updating}, we provide an instantiation MOREC-MOPO, which utilizes MOPO \citep{yu2020mopo} as the policy update component. The uncertainty estimation in line~\ref{mopo_line_uncertainty} is elaborated in Appendix~\ref{app_uncertain_estimation}. In the same way MOREC-MOBILE is the algorithm that the SOTA method MOBILE \citep{sum2023mobile} has MOREC plugged in, which  will be employed in our experiments.

\begin{figure*}[t]
\centering 
\includegraphics[width=1.0 \linewidth]{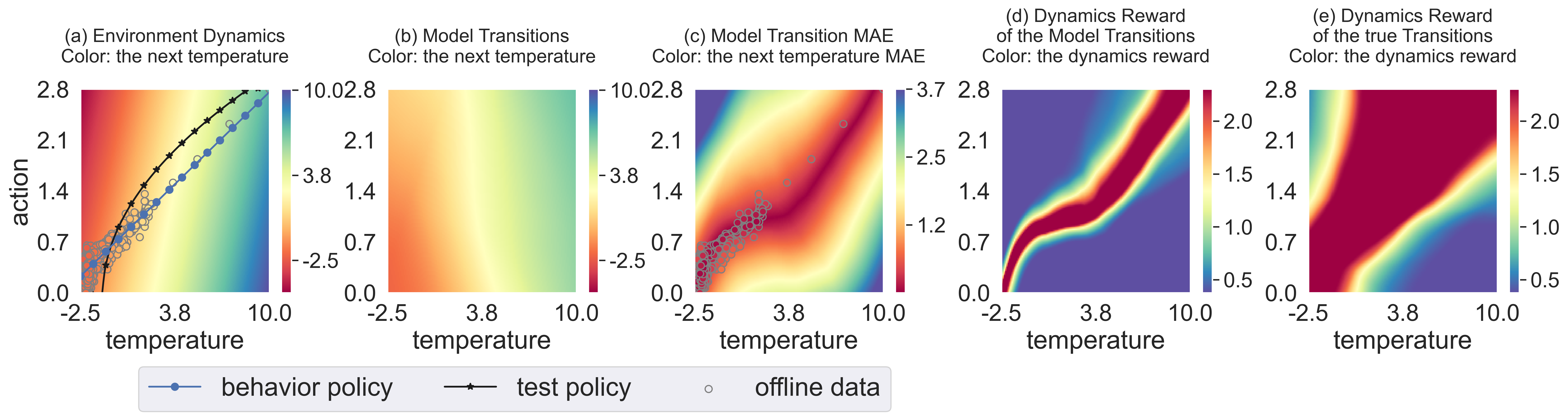}
\caption{Illustration of the dynamics, behavior policy, model MAE, dynamics reward, and test policy in the refrigerator temperature-control task. Panel (a) visualizes the relationship between temperature, action, and the next temperature, mapped to the $x$, $y$, and color axes, respectively. A point at coordinates $(x,y)$, colored by value $c$, indicates that applying action $y$ at temperature $x$ results in a next temperature of $c$. For panels (b)-(e), the interpretation of the color axes is clarified within the respective figure titles. 
}
\label{fig_refrigrator_result}  
\end{figure*}

\subsection{Connection with Adversarial Model Learning}

Adversarial model learning employs the generative adversarial imitation learning framework to learn the dynamics model \citep{shi2019taobao,xu2022model_imitation}. It also uses the objective as Eq.\eqref{eq_original_discriminator_loss}. The equivalence between adversarial imitation learning and inverse reinforcement learning \citep{finn2016connection} highlights the shared principle with MOREC, which explicitly learns the reward model. Both methods benefit from small compounding error \citep{xu2022model_imitation} and exhibit causality-consistent transitions \citep{chen2022galileo}.

However, there are noticeable differences. MOREC employs an ensemble of discriminators as the reward model, which is more stable and generalizable compared to using a single discriminator as in previous adversarial model learning methods \citep{luo2022trasfer}. Additionally, the filtering mechanism allows supervised and adversarial model learning to complement each other. This is particularly advantageous for large training data, as supervised model learning helps MOREC leverage all the available data. In contrast, previous adversarial model learning methods solely rely on generative adversarial learning, which may fail to capture small modes in the data.

%\secspace
\section{Experiments}
%\secspace
In this section, we conduct a series of experiments designed to answer the following questions:
\begin{enumerate}
    \item[\textbf{Q1. }] What is the appearance of the recovered dynamics reward? (\cref{fig_refrigrator_result}, \ref{fig_d4rl_reward_vis}) 
    \item[\textbf{Q2. }] Can the recovered dynamics reward enhance the accuracy of rollouts? (\cref{fig_refrigerator_refine}, \ref{fig_rollout_mae_compare})
    \item[\textbf{Q3. }] Can \ours facilitate learning policies with superior performance? (\cref{fig_refrigerator_policy_vis}, Table~\ref{tab_refrigerator_return}, \ref{tab_d4rl_score}, \ref{tab_neorl_score})
\end{enumerate}

\subsection{Experimental Setup}
In our experiments, we consider a synthetic refrigerator temperature-control task, $12$ locomotion tasks as part of the D4RL benchmark~\citep{fu2020d4rl}, and $9$ locomotion tasks included in the NeoRL benchmark~\citep{qin2022neorl}. Comprehensive experimental details are provided in Appendix~\ref{app_experiment_details}.

\subsection{Evaluation on a Synthetic Task}
\label{sec_RTC_task}
We first consider a synthetic refrigerator temperature-control task where we can delve deeply into the performance of \ours through visualizations. In this task, an agent controls the compressor's power in a refrigerator with the primary aim of maintaining a preset target temperature. The agent's observations are the present temperature, with the action corresponding to the normalized power of the compressor. We depict the dynamics of this system in \cref{fig_refrigrator_result}~(a), which maps temperature, action, and the next temperature to the $x$, $y$, and color axes, respectively. We then use a linear behavior policy to collect the offline dataset (also shown in \cref{fig_refrigrator_result}~(a)) from the environment. % The behavior policy and offline dataset are also shown in \cref{fig_refrigrator_result}~(a). 

We train a dynamics model with the offline data which is shown in \cref{fig_refrigrator_result}~(b). To compare the dynamics model and the true dynamics more clearly, we showcase the mean absolute error (MAE) between the next temperatures output by the true dynamics and the dynamics model in \cref{fig_refrigrator_result} (c). 
A comparative analysis of \cref{fig_refrigrator_result} (a), (b), and (c) suggests the model can align closely with the true dynamics where the offline data is dense, but deviate in data-scarce areas. Notably, the low-error area in \cref{fig_refrigrator_result} (c), denoted in red, mainly corresponds to the regions visited by the behavior policy.

\begin{figure}[t]
  \centering
  
  \begin{minipage}{0.56\textwidth}
    \includegraphics[width=\textwidth]{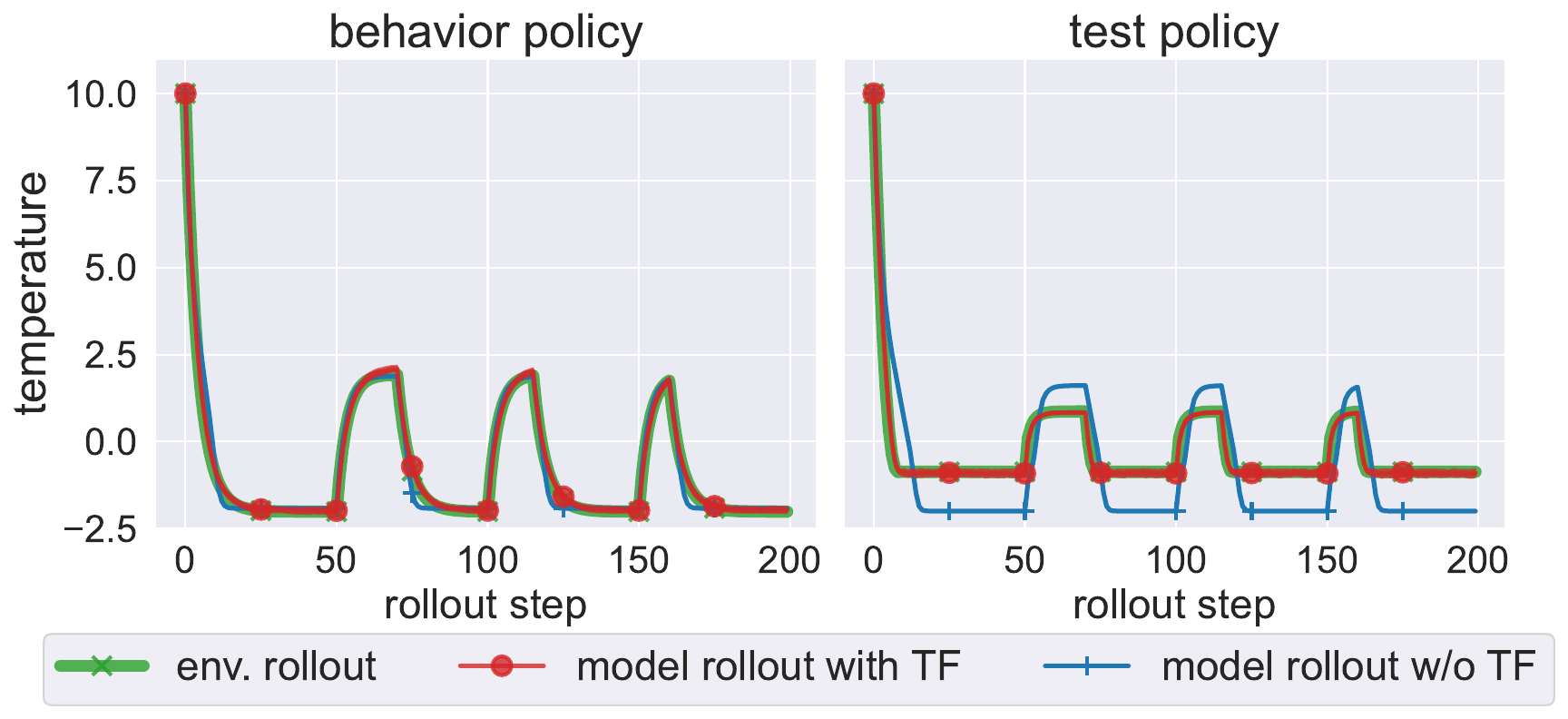}
    \caption{200-step rollouts in model or environment with the behavior policy or test policy (TF: transition filtering).} 
    \label{fig_refrigerator_refine}
  \end{minipage}
  \hspace{1em}
  \begin{minipage}{0.33\textwidth}
    \includegraphics[width=\textwidth]{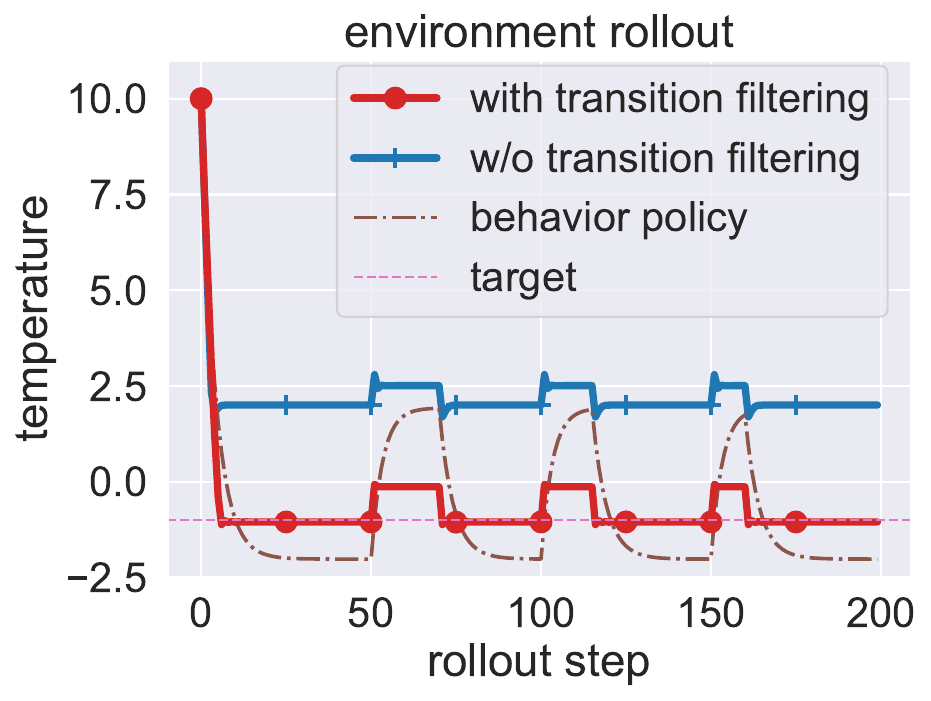}
    \caption{Real environment rollouts of the policies learned with and without transition filtering.}
    \label{fig_refrigerator_policy_vis}
  \end{minipage}
\end{figure}

To answer \textbf{Q1}, we further train a dynamics reward according to Algorithm~\ref{alg_model_reward_learning}. We depict the inferred dynamics rewards of the model transitions and true transitions in \cref{fig_refrigrator_result} (d) and (e), respectively. A noteworthy observation is the strong correlation between the inferred dynamics rewards and the MAE of the next temperatures, as shown in \cref{fig_refrigrator_result} (c) and (d). The dynamics reward aptly allocates higher values to model transitions corresponding to a lower MAE. Moreover, the true transitions generally exhibit higher dynamics rewards than model transitions, with exceptions in areas significantly distant from offline data. This result implies that the learned dynamics reward can accurately identify high-fidelity transitions, i.e., those data that accurately replicate the true dynamics, across a broad range of the state-action space. This suggests a superior generalizability of the dynamics reward compared to the dynamics model.

To answer \textbf{Q2}, we subsequently compare the accuracy of model rollouts generated with and without the proposed transition filtering technique detailed in \eqref{eq_ours_tc_generating}. Here we consider two rollout policies: the behavior policy and another different test policy which is illustrated in \cref{fig_refrigrator_result} (a). \cref{fig_refrigerator_refine} shows the true environment rollout and model rollouts. First, we focus on the behavior policy case, which is shown in the left sub-figure in \cref{fig_refrigerator_refine}. As the dynamics model is trained on the offline dataset collected by the behavior policy, it can give accurate next-state predictions on in-distribution state-action pairs visited by the behavior policy. Thus, both model rollouts with and without transition filtering can perfectly replicate the true environment rollout.

However, in model-based offline RL, our primary concern is the accuracy of model rollouts induced by learning policies, which could deviate from the behavior policy and visit out-of-distribution (OOD) state-action pairs. Thus we turn to the test policy case shown in the right sub-figure in \cref{fig_refrigerator_refine}. As the dynamics model cannot give accurate predictions in a single attempt on OOD state-action pairs visited by the test policy, the model rollout generated without the transition filtering technique deviates from the true one significantly. Nevertheless, when equipped with the transition filtering technique, the generated model rollout still perfectly matches the true rollout. These results clearly demonstrate the effectiveness of utilizing the dynamics reward to generate high-fidelity model rollouts on OOD regions. This further suggests that the learned dynamics reward could exhibit a superior generalization ability than the dynamics model. 

\begin{wraptable}{r}{0.39\textwidth}

% \begin{table}[ht]
 \centering
\small
  \caption{Returns of the learned policies.}
  \label{tab_refrigerator_return}
  \begin{tabular}{lr}
    \toprule
    \textbf{Method} & \textbf{Return} \\
    \midrule
    \textbf{with transition filtering}  & $\mathbf{-78}$ \\
    \textbf{w/o transition filtering}  & $-640$ \\
    \textbf{behavior policy} & $-270$ \\
    \bottomrule
  \end{tabular}
% \end{table}
\end{wraptable}

To answer \textbf{Q3}, we train a policy that aims to control the temperature to $-1^\circ$C with the reward function $r (\text{temperature}_t) = - \vert \text{temperature}_t + 1 \vert$. In particular, we apply PPO~\citep{schulman2017ppo} to learn policies from model rollouts generated with and without the proposed transition filtering technique. The rollouts of the learned policies evaluated in the \emph{true} environment are visualized in \cref{fig_refrigerator_policy_vis}. We observe that the policy learned with transition filtering can effectively control the temperature to $-1^\circ$C. Unfortunately, the policy learned without transition filtering induces a temperature that significantly differs from the target. The detailed returns of the learned policies are reported in Table \ref{tab_refrigerator_return}. We see that the policy learned with transition filtering outperforms that learned without transition filtering by a wide margin. These observations validate that utilizing the dynamics reward can substantially boost the performance of policy learning.

\subsection{Evaluation on the D4RL and NeoRL Benchmark}

\begin{table*}[t]
    \centering
    % \small
    \caption{Normalized average returns in 12 D4RL tasks, averaged over \(5\) seeds. \textit{Solved tasks} denotes the number of the tasks whose scores $\geq 95.0$. The previously best results are underlined.}
    \resizebox{\linewidth}{!}{
    \begin{tabular}{l|cccccccc|cc}
\toprule
\textbf{Task Name}   & \textbf{CQL}  & \textbf{TD3+BC} & \textbf{EDAC}  & \textbf{MOPO} & \textbf{COMBO} & \textbf{TT}   & \textbf{RAMBO} & \textbf{MOBILE} & \textbf{\ours-MOPO} & \textbf{\ours-MOBILE} \\
\midrule
hfctah-rnd          & $31.3$ & $11.0$   & $28.4$  & $38.5$  & $38.8$  & $6.1$  & $39.5$  & $\underline{39.3}$  &    $51.6\pm 0.5$   &    $\mathbf{53.2}\pm \mathbf{1.4}$     \\
hopper-rnd          & $5.3$  & $8.5$    & $25.3$  & $31.7$  & $17.9$  & $6.9$  & $25.4$  & $\underline{31.9}$  &    $\mathbf{32.1}\pm \mathbf{0.1}$ &    $26.6\pm 4.0$     \\
walker-rnd          & $5.4$  & $1.6$    & $16.6$  & $7.4$   & $7.0$   & $5.9$  & $0.0$   & $\underline{17.9}$  &    $\mathbf{23.5}\pm \mathbf{0.7}$ &    $\mathbf{22.8}\pm \mathbf{0.9}$     \\
\midrule
hfctah-med          & $46.9$ & $48.3$   & $65.9$  & $73.0$  & $54.2$  & $46.9$ & $77.9$  & $\underline{74.6}$  &    $\mathbf{82.3}\pm \mathbf{1.1}$ &    $\mathbf{82.1}\pm \mathbf{1.2}$     \\
hopper-med          & $61.9$ & $59.3$   & $101.6$ & $62.8$  & $97.2$  & $67.4$ & $87.0$  & $\underline{106.6}$ &    $107.0\pm 0.3$  &    $\mathbf{108.0}\pm \mathbf{0.5}$    \\
walker-med          & $79.5$ & $83.7$   & $\mathbf{\underline{92.5}}$ & $84.1$  & $81.9$  & $81.3$ & $84.9$  & $87.7$  &    $89.9\pm 0.7$    &    $85.8\pm 0.5$     \\
\midrule
hfctah-med-rep      & $45.3$ & $44.6$   & $61.3$  & $\underline{72.1}$  & $55.1$  & $44.1$ & $68.7$  & $71.7$  &    $\mathbf{76.5}\pm \mathbf{1.2}$   &    $\mathbf{76.4}\pm \mathbf{0.6}$     \\
hopper-med-rep      & $86.3$ & $60.9$   & $101.0$ & $103.5$ & $89.5$  & $99.4$ & $99.5$  & $\underline{103.9}$ &    $\mathbf{105.1}\pm \mathbf{0.4}$  &    $\mathbf{105.5}\pm \mathbf{1.6}$    \\
walker-med-rep      & $76.8$ & $81.8$   & $87.1$  & $85.6$  & $56.0$  & $82.6$ & $89.2$  & $\underline{89.9}$  &    $\mathbf{95.5}\pm \mathbf{2.0}$    &    $\mathbf{95.8}\pm \mathbf{1.8}$     \\
\midrule
hfctah-med-exp      & $95.0$ & $90.7$   & $106.3$ & $90.8$  & $90.0$  & $95.0$ & $95.4$  & $\underline{108.2}$ &    $\mathbf{112.1}\pm \mathbf{1.8}$ &    $\mathbf{110.9}\pm \mathbf{0.56}$   \\
hopper-med-exp      & $96.9$ & $98.0$   & $110.7$ & $81.6$  & $111.1$ & $110.0$ & $88.2$ & $\underline{112.6}$ &    $\mathbf{113.3}\pm \mathbf{0.2}$ &    $111.5\pm 0.4$       \\
walker-med-exp      & $109.1$ & $110.1$ & $114.7$ & $112.9$ & $103.3$ & $101.9$ & $56.7$ & $\mathbf{\underline{115.2}}$  & $\mathbf{115.8}\pm \mathbf{0.8}$  & $\mathbf{115.5}\pm \mathbf{0.9}$  \\
\midrule
Average             & $61.6$ & $58.2$   & $76.0$  & $70.3$  & $66.8$  & $62.3$  & $67.7$  & $\underline{80.0}$  &    $\mathbf{83.7}$   & $82.8$ \\
Solved tasks & $3/12$ & $2/12$ & $5/12$ & $2/12$ & $3/12$ & $4/12$ & $2/12$ & $\underline{5/12}$ & $\mathbf{6/12}$ & $\mathbf{6/12}$ \\
\bottomrule
\end{tabular}
}
    \label{tab_d4rl_score}
\end{table*}
\begin{table*}[t]
    \centering
    \small
    \caption{Normalized average returns in 9 NeoRL tasks, averaged over $5$ seeds. \textit{Solved tasks} denotes the number of the tasks whose scores $\geq 95.0$. The previously best results are underlined.}
    \resizebox{0.9\linewidth}{!}{
    \begin{tabular}{l|cccccc|cc}
\toprule
\textbf{Task Name}                 & \textbf{BC}   & \textbf{CQL}  & \textbf{TD3+BC} & \textbf{EDAC}  & \textbf{MOPO} & \textbf{MOBILE} & \textbf{\ours-MOPO} & \textbf{\ours-MOBILE} \\
\midrule
HalfCheetah-L        & $29.1$  & $38.2$ & $30.0$   & $31.3$  & $40.1$  & $\mathbf{\underline{54.7}}$ & $53.5\pm 0.6$ &  $51.2\pm0.99$\\
Hopper-L             & $15.1$  & $16.0$  & $15.8$    & $18.3$  & $6.2$  & $\underline{17.4}$ & $\mathbf{25.4}\pm\mathbf{1.3}$ & $\mathbf{26.8}\pm \mathbf{1.4}$ \\
Walker2d-L           & $28.5$  & $\underline{44.7}$  & $43.0$    & $40.2$  & $11.6$  & $37.6$ & $\mathbf{65.0}\pm\mathbf{1.3}$ & $56.8\pm 1.9$ \\\midrule
HalfCheetah-M        & $49.0$ & $54.6$ & $52.3$   & $54.9$  & $62.3$  & $\underline{77.8}$ & $84.1\pm 0.5$ &  $\mathbf{86.0}\pm \mathbf{1.3}$\\
Hopper-M             & $51.3$ & $64.5$ & $\underline{70.3}$   & $44.9$  & $1.0$  & $51.1$ & $83.5\pm 3.8$ &  $\mathbf{109.2}\pm \mathbf{0.6}$\\
Walker2d-M           & $48.7$ & $57.3$ & $58.5$   & $57.6$  & $39.9$  & $\underline{62.2}$ & $\mathbf{76.6}\pm\mathbf{1.7}$ &  $71.1 \pm 0.6$\\\midrule
HalfCheetah-H & $71.3$ & $77.4$ & $75.3$   & $81.4$  & $65.9$  & $\underline{83.0}$ & $90.3\pm 2.0$ & $\mathbf{98.5}\pm \mathbf{0.8}$ \\
Hopper-H      & $43.1$ & $76.6$ & $75.3$   & $52.5$  & $11.5$  & $\underline{87.8}$ & $72.8\pm 3.8$ & $\mathbf{108.5}\pm \mathbf{0.3}$ \\
Walker2d-H    & $72.6$ & $75.3$ & $69.6$   & $\underline{75.5}$  & $18.0$  & $74.9$ & $\mathbf{83.0}\pm\mathbf{1.4}$ & $79.3\pm 0.6$ \\\midrule % 79.3 \pm 0.6
Average       & $45.4$ & $56.1$ & $54.5$   & $50.7$  & $28.5$  & $\underline{60.7}$ & $70.3$ & $\mathbf{76.4}$ \\
Solved tasks & $0/9$ & $0/9$ & $0/9$ & $0/9$ & $0/9$ & $0/9$ & $0/9$ & $\mathbf{3/9}$ \\
\bottomrule
\end{tabular}
}
    \label{tab_neorl_score}
\end{table*}
To assess the applicability of \ours in more complex tasks, we validate it on 21 robot locomotion control tasks from the D4RL \citep{fu2020d4rl} and NeoRL \citep{qin2022neorl} benchmarks. Notice that the NeoRL benchmark is more challenging than the D4RL benchmark as the offline datasets in NeoRL are collected from more narrow distributions \citep{qin2022neorl}.
\\
\textbf{Overall Performance. } The normalized average returns are reported in Tables \ref{tab_d4rl_score} and \ref{tab_neorl_score}. We observe that MOREC-MOPO and MOREC-MOBILE outperform prior offline RL methods on $18$ and $17$ of total $21$ tasks, respectively. On both the D4RL and NeoRL benchmarks, MOREC-MOPO and MOREC-MOBILE respectively achieve substantial improvements over MOPO and MOBILE, which clearly demonstrate the benefit from the dynamics reward on policy learning and thus effectively respond to \textbf{Q3}. In the more challenging NeoRL benchmark, MOREC offers a more significant improvement over the existing offline RL methods. In particular, MOREC-MOBILE outperforms the SOTA method MOBILE in terms of the average return by a wide margin of $15.7$, representing an approximate improvement of $25.9 \%$. Furthermore, we also report the number of solved tasks which refer to tasks with normalized returns exceeding $95$. In NeoRL, MOREC-MOBILE is the first method that can solve three tasks while previous methods could not solve any of them. We present the learning curves of both methods in Appendix~\ref{app_additional_performance}.

\begin{figure}[ht]
     \centering
     \begin{subfigure}[b]{0.31\textwidth}
         \centering
         \includegraphics[width=\textwidth]{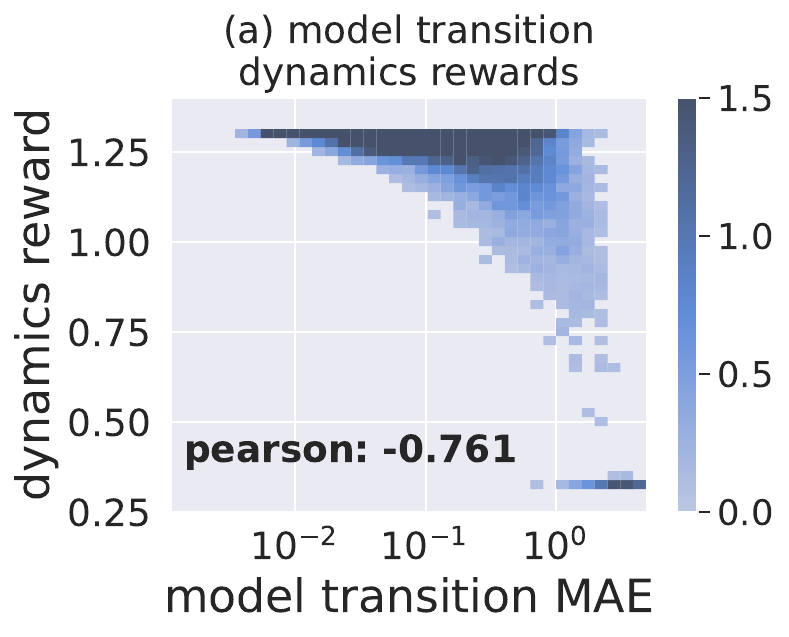}
         \caption{$r^D(s_t,a_t,s_{t+1})$}
         \label{fig_d4rl_reward_vis_a}
     \end{subfigure}
     \hfill
     \begin{subfigure}[b]{0.31\textwidth}
         \centering
         \includegraphics[width=\textwidth]{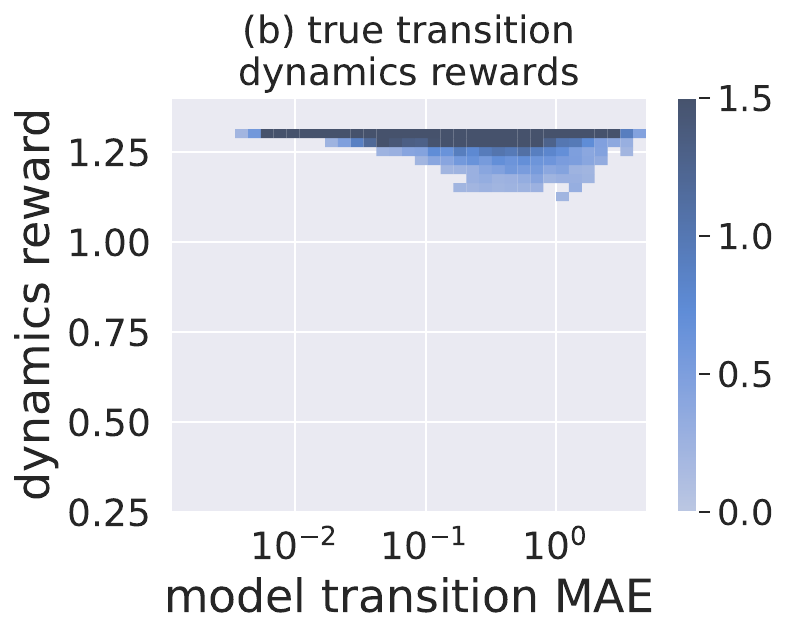}
         \caption{$r^D(s_t,a_t,s_{t+1}^\star)$}
         \label{fig_d4rl_reward_vis_b}
     \end{subfigure}
     \hfill
     \begin{subfigure}[b]{0.35\textwidth}
         \centering
         \includegraphics[width=\textwidth]{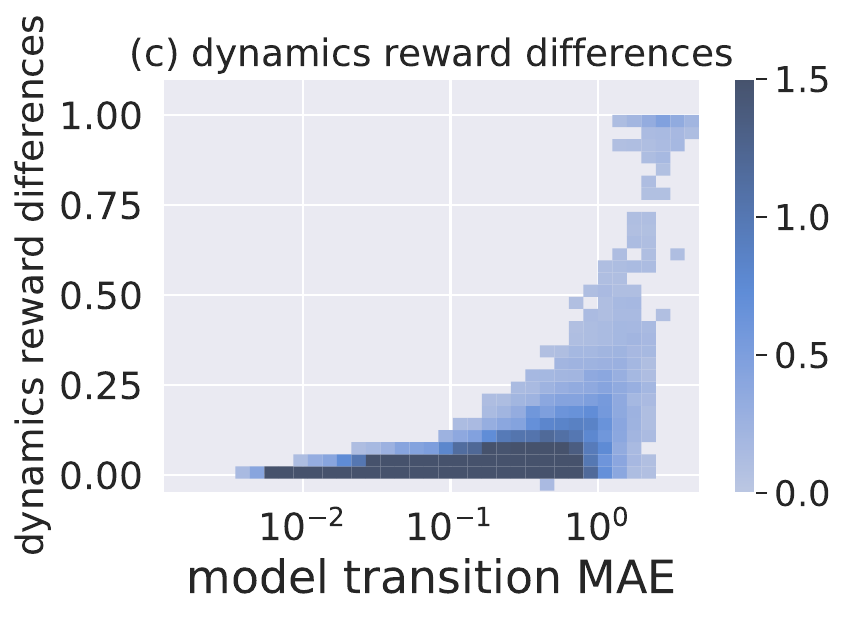}
         \caption{
         $r^D(s_t,a_t,s_{t+1}^\star) - r^D(s_t,a_t,s_{t+1})$
         }
         \label{fig_d4rl_reward_vis_c}
     \end{subfigure}
        \caption{The joint distributions of the model transition MAE, and the dynamics reward $r^D(\cdot)$ for both model transitions and true transitions in the \texttt{walker-med-rep} task. The $x$-axes (\textit{model transition MAE}) all denote the MAE between the true transition and the model transition: $\Vert s_{t+1} - s_{t+1}^\star\Vert_1$.}
        \label{fig_d4rl_reward_vis}
\end{figure}
\begin{figure}[t]
\centering 
\includegraphics[width=1.0 \linewidth]{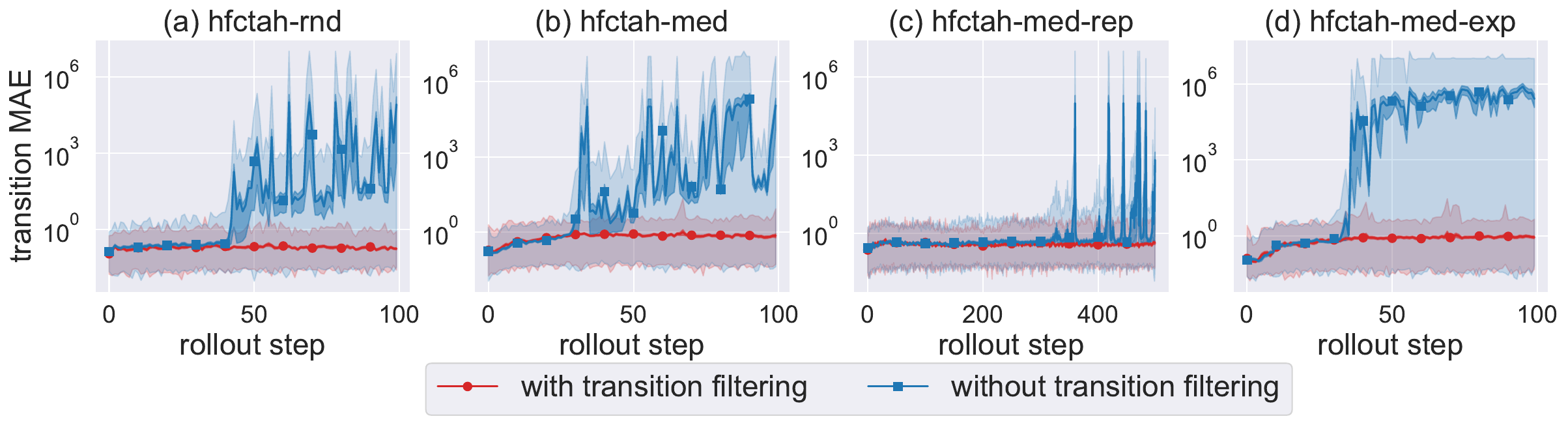}
\caption{The MAE in terms of rollout steps with or without transition filtering technique using the dynamics reward.} 
\label{fig_rollout_mae_compare}  
\end{figure}

\textbf{Performance of the Dynamics Reward. } To answer \textbf{Q1}, we design experiments to evaluate whether the dynamics reward learned by Algorithm \ref{alg_model_reward_learning} can assign transitions proper rewards that are well-aligned with their accuracy. In particular, we choose the \texttt{walker-med-rep} task and apply a policy learned by behaviour cloning in the learned dynamics model to collect model rollouts $\{ (s_t, a_t, s_{t+1}) \}_t$ with $500$ trajectories. Then we record the dynamics reward $r^{D} (s_t, a_t, s_{t+1})$ and MAE $\| s_{t+1} - s^{\star}_{t+1} \|_{1}$ on the collected model rollouts, where $s^\star_{t+1}$ is the ground-truth next state. The joint distribution of $\{ ( \| s_{t+1} - s^\star_{t+1} \|_1, r^{D} (s_t, a_t, s_{t+1})) \}_t$ is plotted in \cref{fig_d4rl_reward_vis_a}. We observe a strong negative correlation between the dynamics reward and the MAE. The detailed Pearson correlation between the dynamics reward and the model transition MAE is $-0.761$, which also denotes a strong negative correlation between both variables. Besides, we remark that the RMSE on the offline dataset is $0.63$ (refer to Table~\ref{tab_d4rl_model_holdout_loss} in Appendix \ref{subsec:the_performance_of_the_learned_model}), and thus the transitions with extremely large MAE (e.g., $3.0$) could be regarded as out of distribution data. Even on such OOD transitions, the learned dynamics reward still can assign proper rewards that are consistent with their accuracy, demonstrating its generalization ability. These results clearly demonstrate that the learned dynamics reward is an excellent indicator of the fidelities of transitions.

Moreover, we calculate dynamics rewards on the \emph{true} transitions $r^{D} (s_t, a_t, s^\star_{t+1})$. The joint distributions of $\{ (\| s_{t+1} - s^\star_{t+1} \|_1, r^{D} (s_t, a_t, s^\star_{t+1})) \}$ and $\{ (\| s_{t+1} - s^\star_{t+1} \|_1, r^{D} (s_t, a_t, s^\star_{t+1}) - r^{D} (s_t, a_t, s_{t+1})) \}$ are reported in \cref{fig_d4rl_reward_vis_b} and \cref{fig_d4rl_reward_vis_c}, respectively. We see that the learned dynamics reward can accurately assign high values to most true transitions. Note that the transitions $(s_t, a_t, s_{t+1})$ and $(s_t, a_t, s^\star_{t+1})$ only differ in the next state. Nevertheless, the dynamics reward gives totally different values, which suggests that it accurately pays attention to the next state transition. In summary, we empirically verify that the dynamics reward learned by Algorithm \ref{alg_model_reward_learning} is capable of assigning transitions proper rewards that are well-aligned with their accuracy, and thus identifying high-fidelity transitions from all candidates. More visualization of the dynamics model differences are presented in Appendix~\ref{app_additional_reward_difference}.            
\\
\textbf{Effectiveness of the Transition Filtering Technique. } To answer \textbf{Q2}, we conduct an ablation study to verify the effectiveness of utilizing the learned dynamics reward to filter transitions, which is detailed in \eqref{eq_ours_tc_generating}. Here we choose the \texttt{hfctah} tasks where there is no terminal state, which allows the rollout horizon to reach the preset value. We rollout the policy learned by MOREC-MOPO in the learned dynamics model. In particular, we consider two rollout processes: one with the transition filtering technique and the other one without the technique. We respectively take these two rollout processes to collect $100$ trajectories and show the MAE on the collected trajectories in \cref{fig_rollout_mae_compare}. On the one hand, when equipped with the transition filtering technique, the generated model rollouts always keep a relatively small MAE as the rollout step increases. On the other hand, without the transition filtering technique, the MAE on the generated model rollouts blows up as the rollout step increases. This phenomenon clearly demonstrates that utilizing the dynamics reward to filter transitions can effectively reduce compounding errors, and thus provide high-fidelity model rollouts for policy learning. An extended ablation study can be found in Appendix~\ref{app_ablation_studies}.

\section{Conclusion}
In this paper, we propose the \ours method based on the idea of reward-consistent models. \ours learns a generalizable dynamics reward which is then used as a transition filter in most offline MBRL methods. We evaluate \ours through extensive experiments. \ours outperforms prior methods in $18$ out of $21$ tasks from the offline RL benchmarks. We empirically validate that the recovered dynamics reward is well-aligned with the accuracy of model transitions even on OOD regions, demonstrating its generalization ability. Consequently, utilizing such a generalizable dynamics reward to filter transitions can effectively reduce compounding errors.       

However, despite its strengths, there is still significant scope for improving \ours. Its performance is largely contingent on the accuracy of the dynamics reward recovered by IRL, which presents an interesting future direction: the development of advanced IRL methods tailored for dynamics reward learning. Additionally, it would be valuable to establish theoretical underpinnings for the dynamics reward from the perspective of imitation learning \citep{rajaraman2020fundamental, xu2022model_imitation,  xu2022understanding, li2023theoretical}.  
\section*{Acknowledgement}
We would like to thank Dr. Zuolin Tu for his constructive advice to improve this paper.

\bibliography{morec_bib}

\begin{thebibliography}{58}
\providecommand{\natexlab}[1]{#1}
\providecommand{\url}[1]{\texttt{#1}}
\expandafter\ifx\csname urlstyle\endcsname\relax
  \providecommand{\doi}[1]{doi: #1}\else
  \providecommand{\doi}{doi: \begingroup \urlstyle{rm}\Url}\fi

\bibitem[Abbeel \& Ng(2004)Abbeel and Ng]{ng2004apprentice}
Pieter Abbeel and Andrew~Y. Ng.
\newblock Apprenticeship learning via inverse reinforcement learning.
\newblock In \emph{Proceedings of the International Conference on Machine
  Learning}, Alberta, Canada, 2004.

\bibitem[Amin \& Singh(2016)Amin and Singh]{amin2016resolving}
Kareem Amin and Satinder Singh.
\newblock Towards resolving unidentifiability in inverse reinforcement
  learning.
\newblock \emph{CoRR}, abs/1601.06569, 2016.

\bibitem[An et~al.(2021)An, Moon, Kim, and Song]{an2021uncertainty}
Gaon An, Seungyong Moon, Jang-Hyun Kim, and Hyun~Oh Song.
\newblock Uncertainty-based offline reinforcement learning with diversified
  {Q}-ensemble.
\newblock In \emph{Advances in Neural Information Processing Systems}, virtual,
  2021.

\bibitem[Bai et~al.(2022)Bai, Wang, Yang, Deng, Garg, Liu, and
  Wang]{bai2022pessimistic}
Chenjia Bai, Lingxiao Wang, Zhuoran Yang, Zhi{-}Hong Deng, Animesh Garg, Peng
  Liu, and Zhaoran Wang.
\newblock Pessimistic bootstrapping for uncertainty-driven offline
  reinforcement learning.
\newblock In \emph{Proceedings of the International Conference on Learning
  Representations}, virtual, 2022.

\bibitem[Boyd \& Vandenberghe(2004)Boyd and Vandenberghe]{boyd2004convex}
Stephen~P Boyd and Lieven Vandenberghe.
\newblock \emph{Convex optimization}.
\newblock Cambridge University Press, 2004.

\bibitem[Chen et~al.(2021)Chen, Yu, Li, Luo, Qin, Shang, and Ye]{chen2021maple}
Xiong{-}Hui Chen, Yang Yu, Qingyang Li, Fan{-}Ming Luo, Zhiwei~(Tony) Qin,
  Wenjie Shang, and Jieping Ye.
\newblock Offline model-based adaptable policy learning.
\newblock In \emph{Advances in Neural Information Processing Systems}, virtual,
  2021.

\bibitem[Chen et~al.(2023)Chen, Yu, Zhu, Yu, Chen, Wang, Wu, Wu, Qin, Ding, and
  Huang]{chen2022galileo}
Xiong-Hui Chen, Yang Yu, Zheng-Mao Zhu, Zhihua Yu, Zhenjun Chen, Chenghe Wang,
  Yinan Wu, Hongqiu Wu, Rong-Jun Qin, Ruijin Ding, and Fangsheng Huang.
\newblock Adversarial counterfactual environment model learning.
\newblock In \emph{Advances in Neural Information Processing Systems}, New
  Orleans, LA, 2023.

\bibitem[Finn et~al.(2016{\natexlab{a}})Finn, Christiano, Abbeel, and
  Levine]{finn2016connection}
Chelsea Finn, Paul~F. Christiano, Pieter Abbeel, and Sergey Levine.
\newblock A connection between generative adversarial networks, inverse
  reinforcement learning, and energy-based models.
\newblock \emph{CoRR}, abs/1611.03852, 2016{\natexlab{a}}.

\bibitem[Finn et~al.(2016{\natexlab{b}})Finn, Levine, and Abbeel]{finn2016gcl}
Chelsea Finn, Sergey Levine, and Pieter Abbeel.
\newblock Guided cost learning: Deep inverse optimal control via policy
  optimization.
\newblock In \emph{Proceedings of the International Conference on Machine
  Learning}, New York City, NY, 2016{\natexlab{b}}.

\bibitem[Fu et~al.(2017)Fu, Luo, and Levine]{fu2017airl}
Justin Fu, Katie Luo, and Sergey Levine.
\newblock Learning robust rewards with adversarial inverse reinforcement
  learning.
\newblock \emph{CoRR}, abs/1710.11248, 2017.

\bibitem[Fu et~al.(2020)Fu, Kumar, Nachum, Tucker, and Levine]{fu2020d4rl}
Justin Fu, Aviral Kumar, Ofir Nachum, George Tucker, and Sergey Levine.
\newblock {D4RL}: Datasets for deep data-driven reinforcement learning.
\newblock \emph{CoRR}, abs/2004.07219, 2020.

\bibitem[Fujimoto \& Gu(2021)Fujimoto and Gu]{fujimoto2021minimalist}
Scott Fujimoto and Shixiang~Shane Gu.
\newblock A minimalist approach to offline reinforcement learning.
\newblock In \emph{Advances in Neural Information Processing Systems}, virtual,
  2021.

\bibitem[Fujimoto et~al.(2019)Fujimoto, Meger, and Precup]{fujimoto2019off}
Scott Fujimoto, David Meger, and Doina Precup.
\newblock Off-policy deep reinforcement learning without exploration.
\newblock In \emph{Proceedings of the International Conference on Machine
  Learning}, Long Beach, CA, 2019.

\bibitem[Geng et~al.(2020)Geng, Nassif, Manzanares, Reppen, and
  Sircar]{geng2020pqr}
Sinong Geng, Houssam Nassif, Carlos~A. Manzanares, A.~Max Reppen, and Ronnie
  Sircar.
\newblock Deep {PQR}: Solving inverse reinforcement learning using anchor
  actions.
\newblock In \emph{Proceedings of the International Conference on Machine
  Learning}, virtual, 2020.

\bibitem[Ghasemipour et~al.(2019)Ghasemipour, Zemel, and
  Gu]{ghasemipour2019divergence}
Seyed Kamyar~Seyed Ghasemipour, Richard~S. Zemel, and Shixiang Gu.
\newblock A divergence minimization perspective on imitation learning methods.
\newblock In \emph{Proceedings of the Conference on Robot Learning}, Osaka,
  Japan, 2019.

\bibitem[Hazan(2016)]{Hazan16introduction-to-oco}
Elad Hazan.
\newblock Introduction to online convex optimization.
\newblock \emph{Foundations and Trends in Optimization}, 2\penalty0
  (3-4):\penalty0 157--325, 2016.

\bibitem[Ho \& Ermon(2016)Ho and Ermon]{ho2016gail}
Jonathan Ho and Stefano Ermon.
\newblock Generative adversarial imitation learning.
\newblock In \emph{Advances in Neural Information Processing Systems},
  Barcelona, Spain, 2016.

\bibitem[Jacq et~al.(2019)Jacq, Geist, Paiva, and
  Pietquin]{jacq2019learnfromlearner}
Alexis Jacq, Matthieu Geist, Ana Paiva, and Olivier Pietquin.
\newblock Learning from a learner.
\newblock In \emph{Proceedings of the International Conference on Machine
  Learning}, Long Beach, CA, 2019.

\bibitem[Janner et~al.(2019)Janner, Fu, Zhang, and Levine]{michael2019mbpo}
Michael Janner, Justin Fu, Marvin Zhang, and Sergey Levine.
\newblock When to trust your model: Model-based policy optimization.
\newblock In \emph{Advances in Neural Information Processing Systems},
  Vancouver, Canada, 2019.

\bibitem[Janner et~al.(2021)Janner, Li, and Levine]{janner2021offline}
Michael Janner, Qiyang Li, and Sergey Levine.
\newblock Offline reinforcement learning as one big sequence modeling problem.
\newblock In \emph{Advances in Neural Information Processing Systems}, virtual,
  2021.

\bibitem[Kidambi et~al.(2020)Kidambi, Rajeswaran, Netrapalli, and
  Joachims]{lar2020morel}
Rahul Kidambi, Aravind Rajeswaran, Praneeth Netrapalli, and Thorsten Joachims.
\newblock {MOR}e{L}: Model-based offline reinforcement learning.
\newblock In \emph{Advances in Neural Information Processing Systems}, virtual,
  2020.

\bibitem[Kingma \& Ba(2015)Kingma and Ba]{kingma2015adam}
Diederik~P. Kingma and Jimmy Ba.
\newblock Adam: {A} method for stochastic optimization.
\newblock In \emph{Proceedings of the International Conference on Learning
  Representations}, San Diego, CA, 2015.

\bibitem[Kostrikov et~al.(2019)Kostrikov, Agrawal, Dwibedi, Levine, and
  Tompson]{kostrikov2019dac}
Ilya Kostrikov, Kumar~Krishna Agrawal, Debidatta Dwibedi, Sergey Levine, and
  Jonathan Tompson.
\newblock Discriminator-actor-critic: Addressing sample inefficiency and reward
  bias in adversarial imitation learning.
\newblock In \emph{Proceedings of the International Conference on Learning
  Representations}, New Orleans, LA, 2019.

\bibitem[Kostrikov et~al.(2020)Kostrikov, Nachum, and
  Tompson]{kostrikov2020valuedice}
Ilya Kostrikov, Ofir Nachum, and Jonathan Tompson.
\newblock Imitation learning via off-policy distribution matching.
\newblock In \emph{Proceedings of the International Conference on Learning
  Representations}, Addis Ababa, Ethiopia, 2020.

\bibitem[Kostrikov et~al.(2021)Kostrikov, Fergus, Tompson, and
  Nachum]{kostrikov2021offline}
Ilya Kostrikov, Rob Fergus, Jonathan Tompson, and Ofir Nachum.
\newblock Offline reinforcement learning with fisher divergence critic
  regularization.
\newblock In \emph{Proceedings of the International Conference on Machine
  Learning}, virtual, 2021.

\bibitem[Kumar et~al.(2019)Kumar, Fu, Soh, Tucker, and
  Levine]{kumar2019stabilizing}
Aviral Kumar, Justin Fu, Matthew Soh, George Tucker, and Sergey Levine.
\newblock Stabilizing off-policy {Q}-learning via bootstrapping error
  reduction.
\newblock In \emph{Advances in Neural Information Processing Systems},
  Vancouver, Canada, 2019.

\bibitem[Kumar et~al.(2020)Kumar, Zhou, Tucker, and
  Levine]{kumar2020conservative}
Aviral Kumar, Aurick Zhou, George Tucker, and Sergey Levine.
\newblock Conservative {Q}-learning for offline reinforcement learning.
\newblock In \emph{Advances in Neural Information Processing Systems}, virtual,
  2020.

\bibitem[Levine et~al.(2020)Levine, Kumar, Tucker, and Fu]{levine2020offline}
Sergey Levine, Aviral Kumar, George Tucker, and Justin Fu.
\newblock Offline reinforcement learning: Tutorial, review, and perspectives on
  open problems.
\newblock \emph{CoRR}, abs/2005.01643, 2020.

\bibitem[Li et~al.(2022)Li, Xu, Yu, and Luo]{li2022rethinking}
Ziniu Li, Tian Xu, Yang Yu, and Zhi-Quan Luo.
\newblock Rethinking {V}alue{D}ice: Does it really improve performance?
\newblock \emph{CoRR}, abs/2202.02468, 2022.

\bibitem[Li et~al.(2023)Li, Xu, Yu, and Luo]{li2023theoretical}
Ziniu Li, Tian Xu, Yang Yu, and Zhi{-}Quan Luo.
\newblock Theoretical analysis of offline imitation with supplementary dataset.
\newblock \emph{CoRR}, abs/2301.11687, 2023.

\bibitem[Lin et~al.(2020)Lin, Jin, and Jordan]{lin2020gradient}
Tianyi Lin, Chi Jin, and Michael Jordan.
\newblock On gradient descent ascent for nonconvex-concave minimax problems.
\newblock In \emph{Proceedings of the International Conference on Machine
  Learning}, virtual, 2020.

\bibitem[Luo et~al.(2022{\natexlab{a}})Luo, Cao, and Yu]{luo2022darl}
Fan{-}Ming Luo, Xingchen Cao, and Yang Yu.
\newblock Transferable reward learning by dynamics-agnostic discriminator
  ensemble.
\newblock \emph{CoRR}, abs/2206.00238, 2022{\natexlab{a}}.

\bibitem[Luo et~al.(2022{\natexlab{b}})Luo, Cao, and Yu]{luo2022trasfer}
Fan-Ming Luo, Xingchen Cao, and Yang Yu.
\newblock Transferable reward learning by dynamics-agnostic discriminator
  ensemble.
\newblock \emph{CoRR}, abs/2206.00238, 2022{\natexlab{b}}.

\bibitem[Luo et~al.(2023)Luo, Xu, Lai, Chen, Zhang, and Yu]{luo2022survey}
Fan{-}Ming Luo, Tian Xu, Hang Lai, Xiong{-}Hui Chen, Weinan Zhang, and Yang Yu.
\newblock A survey on model-based reinforcement learning.
\newblock \emph{SCIENCE CHINA Information Sciences}, 2023.

\bibitem[Moerland et~al.(2023)Moerland, Broekens, Plaat, and
  Jonker]{moerland2023mbrlsurvey}
Thomas~M. Moerland, Joost Broekens, Aske Plaat, and Catholijn~M. Jonker.
\newblock Model-based reinforcement learning: {A} survey.
\newblock \emph{Foundations and Trends in Machine Learning}, 16\penalty0
  (1):\penalty0 1--118, 2023.

\bibitem[Nesterov(2003)]{nesterov2003introductory}
Yurii Nesterov.
\newblock \emph{Introductory lectures on convex optimization: A basic course},
  volume~87.
\newblock Springer Science \& Business Media, 2003.

\bibitem[Ng \& Russell(2000)Ng and Russell]{ng2000irl}
Andrew~Y. Ng and Stuart Russell.
\newblock Algorithms for inverse reinforcement learning.
\newblock In \emph{Proceedings of the International Conference on Machine
  Learning}, Stanford, CA, 2000.

\bibitem[Ni et~al.(2020)Ni, Sikchi, Wang, Gupta, Lee, and
  Eysenbach]{ni2020firl}
Tianwei Ni, Harshit~S. Sikchi, Yufei Wang, Tejus Gupta, Lisa Lee, and Ben
  Eysenbach.
\newblock $f$-{IRL}: Inverse reinforcement learning via state marginal
  matching.
\newblock In \emph{Proceedings of the Conference on Robot Learning}, virtual,
  2020.

\bibitem[Peng et~al.(2019)Peng, Kanazawa, Toyer, Abbeel, and
  Levine]{peng2019vdb}
Xue~Bin Peng, Angjoo Kanazawa, Sam Toyer, Pieter Abbeel, and Sergey Levine.
\newblock Variational discriminator bottleneck: Improving imitation learning,
  inverse {RL}, and {GAN}s by constraining information flow.
\newblock In \emph{Proceedings of the International Conference on Learning
  Representations}, New Orleans, LA, 2019.

\bibitem[Qin et~al.(2022)Qin, Zhang, Gao, Chen, Li, Zhang, and
  Yu]{qin2022neorl}
Rongjun Qin, Xingyuan Zhang, Songyi Gao, Xiong{-}Hui Chen, Zewen Li, Weinan
  Zhang, and Yang Yu.
\newblock Neo{RL}: {A} near real-world benchmark for offline reinforcement
  learning.
\newblock In \emph{Advances in Neural Information Processing Systems}, New
  Orleans, LA, 2022.

\bibitem[Rajaraman et~al.(2020)Rajaraman, Yang, Jiao, and
  Ramchandran]{rajaraman2020fundamental}
Nived Rajaraman, Lin~F. Yang, Jiantao Jiao, and Kannan Ramchandran.
\newblock Toward the fundamental limits of imitation learning.
\newblock In \emph{Advances in Neural Information Processing Systems}, virtual,
  2020.

\bibitem[Ratliff et~al.(2006)Ratliff, Bagnell, and
  Zinkevich]{ratliff2006maximum}
Nathan~D Ratliff, J~Andrew Bagnell, and Martin~A Zinkevich.
\newblock Maximum margin planning.
\newblock In \emph{Proceedings of the International Conference on Machine
  Learning}, Pittsburgh, PA, 2006.

\bibitem[Rigter et~al.(2022)Rigter, Lacerda, and Hawes]{rigter2022rambo}
Marc Rigter, Bruno Lacerda, and Nick Hawes.
\newblock {RAMBO-RL}: {R}obust adversarial model-based offline reinforcement
  learning.
\newblock In \emph{Advances in Neural Information Processing Systems}, New
  Orleans, LA, 2022.

\bibitem[Schulman et~al.(2017)Schulman, Wolski, Dhariwal, Radford, and
  Klimov]{schulman2017ppo}
John Schulman, Filip Wolski, Prafulla Dhariwal, Alec Radford, and Oleg Klimov.
\newblock Proximal policy optimization algorithms.
\newblock \emph{CoRR}, abs/1707.06347, 2017.

\bibitem[Shang et~al.(2021)Shang, Li, Qin, Yu, Meng, and Ye]{shang2021mlj}
Wenjie Shang, Qingyang Li, Zhiwei Qin, Yang Yu, Yiping Meng, and Jieping Ye.
\newblock Partially observable environment estimation with uplift inference for
  reinforcement learning based recommendation.
\newblock \emph{Machine Learning}, 110\penalty0 (9):\penalty0 2603--2640, 2021.

\bibitem[Shi et~al.(2019)Shi, Yu, Da, Chen, and Zeng]{shi2019taobao}
Jing{-}Cheng Shi, Yang Yu, Qing Da, Shi{-}Yong Chen, and Anxiang Zeng.
\newblock Virtual-taobao: {V}irtualizing real-world online retail environment
  for reinforcement learning.
\newblock In \emph{Proceedings of the {AAAI} Conference on Artificial
  Intelligence}, Honolulu, HI, 2019.

\bibitem[Sun et~al.(2021)Sun, Mahajan, Hofmann, and Whiteson]{sun2021softdice}
Mingfei Sun, Anuj Mahajan, Katja Hofmann, and Shimon Whiteson.
\newblock Soft{DICE} for imitation learning: Rethinking off-policy distribution
  matching.
\newblock \emph{CoRR}, abs/2106.03155, 2021.

\bibitem[Sun et~al.(2023)Sun, Zhang, Jia, Lin, Ye, and Yu]{sum2023mobile}
Yihao Sun, Jiaji Zhang, Chengxing Jia, Haoxin Lin, Junyin Ye, and Yang Yu.
\newblock Model-{B}ellman inconsistency for model-based offline reinforcement
  learning.
\newblock In \emph{Proceedings of the International Conference on Machine
  Learning}, Honolulu, HI, 2023.

\bibitem[Sutton \& Barto(1998)Sutton and Barto]{sutton1998rl}
Richard~S. Sutton and Andrew~G. Barto.
\newblock \emph{Reinforcement learning: {A}n introduction}.
\newblock MIT Press, 1998.

\bibitem[Sutton et~al.(1999)Sutton, McAllester, Singh, and
  Mansour]{sutton1999policy}
Richard~S. Sutton, David~A. McAllester, Satinder Singh, and Yishay Mansour.
\newblock Policy gradient methods for reinforcement learning with function
  approximation.
\newblock In \emph{Advances in Neural Information Processing Systems}, Denver,
  CO, 1999.

\bibitem[Swamy et~al.(2021)Swamy, Choudhury, Bagnell, and Wu]{swamy2021moments}
Gokul Swamy, Sanjiban Choudhury, J~Andrew Bagnell, and Steven Wu.
\newblock Of moments and matching: A game-theoretic framework for closing the
  imitation gap.
\newblock In \emph{Proceeding of the International Conference on Machine
  Learning}, virtual, 2021.

\bibitem[Vaswani et~al.(2017)Vaswani, Shazeer, Parmar, Uszkoreit, Jones, Gomez,
  Kaiser, and Polosukhin]{vaswani2017transformer}
Ashish Vaswani, Noam Shazeer, Niki Parmar, Jakob Uszkoreit, Llion Jones,
  Aidan~N. Gomez, Lukasz Kaiser, and Illia Polosukhin.
\newblock Attention is all you need.
\newblock In \emph{Advances in Neural Information Processing Systems}, Long
  Beach, CA, 2017.

\bibitem[Xu et~al.(2022{\natexlab{a}})Xu, Li, and Yu]{xu2022model_imitation}
Tian Xu, Ziniu Li, and Yang Yu.
\newblock Error bounds of imitating policies and environments for reinforcement
  learning.
\newblock \emph{{IEEE} Transactions on Pattern Analysis and Machine
  Intelligence}, 44\penalty0 (10):\penalty0 6968--6980, 2022{\natexlab{a}}.

\bibitem[Xu et~al.(2022{\natexlab{b}})Xu, Li, Yu, and Luo]{xu2022understanding}
Tian Xu, Ziniu Li, Yang Yu, and Zhi{-}Quan Luo.
\newblock Understanding adversarial imitation learning in small sample regime:
  {A} stage-coupled analysis.
\newblock \emph{CoRR}, abs/2208.01899, 2022{\natexlab{b}}.

\bibitem[Xu et~al.(2023)Xu, Li, Yu, and Luo]{xu2023provably}
Tian Xu, Ziniu Li, Yang Yu, and Zhi{-}Quan Luo.
\newblock Provably efficient adversarial imitation learning with unknown
  transitions.
\newblock In \emph{Proceedings of the Conference on Uncertainty in Artificial
  Intelligence}, Pittsburgh, PA, 2023.

\bibitem[Yu et~al.(2020)Yu, Thomas, Yu, Ermon, Zou, Levine, Finn, and
  Ma]{yu2020mopo}
Tianhe Yu, Garrett Thomas, Lantao Yu, Stefano Ermon, James~Y. Zou, Sergey
  Levine, Chelsea Finn, and Tengyu Ma.
\newblock {MOPO}: Model-based offline policy optimization.
\newblock In \emph{Advances in Neural Information Processing Systems}, virtual,
  2020.

\bibitem[Yu et~al.(2021)Yu, Kumar, Rafailov, Rajeswaran, Levine, and
  Finn]{yu2021combo}
Tianhe Yu, Aviral Kumar, Rafael Rafailov, Aravind Rajeswaran, Sergey Levine,
  and Chelsea Finn.
\newblock {COMBO}: Conservative offline model-based policy optimization.
\newblock In \emph{Advances in Neural Information Processing Systems}, virtual,
  2021.

\bibitem[Ziebart et~al.(2008)Ziebart, Maas, Bagnell, and
  Dey]{ziebart2008maxent}
Brian~D. Ziebart, Andrew~L. Maas, J.~Andrew Bagnell, and Anind~K. Dey.
\newblock Maximum entropy inverse reinforcement learning.
\newblock In \emph{Proceedings of the {AAAI} Conference on Artificial
  Intelligence}, Chicago, IL, 2008.

\end{thebibliography}
\bibliographystyle{iclr}\
\clearpage
\appendix

\tableofcontents

\newpage
\section{Extended Related Work}

\noindent{\textbf{Offline RL.}} Offline RL studies the methodologies that enable the agent to directly learn an effective policy from an offline experience dataset without any interaction with the environment dynamics. The major challenge of offline RL arises from the discrepancy between the offline experience dataset and the behavior policy's visitation, resulting in extrapolation errors~\citep{fujimoto2019off, kumar2019stabilizing}. Model-free offline RL algorithms~\citep{fujimoto2019off, fujimoto2021minimalist, kostrikov2021offline, kumar2019stabilizing, kumar2020conservative, bai2022pessimistic} incorporate conservatism into policy or Q-function of online RL algorithms to tackle with extrapolation error.

The challenge arises from the discrepancy between the offline experience dataset and the behavior policy's visitation. This mismatch can lead to a poor estimation of the Q-function of the behavior policy, plagued by extrapolation errors~\citep{fujimoto2019off, kumar2019stabilizing}. Consequently, online off-policy RL algorithms fail to be applied in offline RL directly.

To tackle extrapolation error, conservatism is introduced to offline RL algorithms as a common paradigm. Model-free offline RL algorithms incorporate conservatism or regularization into online RL algorithms by preventing the behavior policy from acting in out-of-support regions~\citep{fujimoto2019off, fujimoto2021minimalist, kostrikov2021offline} or by learning a conservative Q-function for out-of-distribution (OOD) visitation of the behavior policy~\citep{kumar2019stabilizing, kumar2020conservative, bai2022pessimistic}, without learning dynamics models.

\noindent{\textbf{Offline MBRL.}} Offline MBRL algorithms~\citep{yu2020mopo, lar2020morel, chen2021maple, yu2021combo} leverage a dynamics model derived from offline data to enhance the efficiency of offline RL. Benefiting from the additional synthetic data generated by the learned dynamics model, model-based algorithms are more likely to have the potential to generalize in the states out of distribution and perform better in new tasks. However, the learned dynamics model also suffers from errors inevitably for the limited experience dataset~\citep{xu2022model_imitation, michael2019mbpo}. Thus, some works~\citep{yu2020mopo, sum2023mobile,lar2020morel, yu2021combo} also incorporate conservatism into offline MBRL.
For example, some methods~\citep{yu2020mopo,bai2022pessimistic,sum2023mobile} apply uncertainty estimation and trust states with low uncertainty, while some methods~\citep{lar2020morel, yu2021combo} try to limit the behavior policy to acting surrounding the experience dataset. However, despite these efforts, the performance of these techniques is largely dependent on the accuracy of the dynamics model.

\noindent{\textbf{Inverse Reinforcement Learning (IRL)}}
IRL~\citep{ng2000irl,ni2020firl, ghasemipour2019divergence, swamy2021moments} is a process that tackles MDPs devoid of reward functions. The aim of IRL is to deduce these reward functions from a series of expert demonstrations. Apprenticeship learning~\citep{ng2004apprentice} trains reward by maximizing an evaluation margin between the expert and the policy. In MaxEnt IRL algorithm, the reward is modeled as a maximum likelihood problem by introducing a maximum entropy objective~\citep{ziebart2008maxent}. GCL~\citep{finn2016gcl} extends MaxEnt IRL to high-dimensional state-action space. Recently, as a special variant of IRL, adversarial imitation learning (AIL)~\citep{ho2016gail,finn2016connection,fu2017airl,kostrikov2019dac,kostrikov2020valuedice,sun2021softdice, li2022rethinking, xu2022understanding} learns policy and reward in a full adversarial manner.

\noindent{\textbf{Generalizable reward learning}}~\citep{fu2017airl}, which aims at learning reward functions robust to the dynamics changes, has attracted lots of interest recently. Generalizable reward learning enables the learned reward to be reusable in a dynamics-mismatch environment, thus largely extending the scope of IRL applications. Prior works have shown such a generalizable reward can be recovered under some assumptions~\citep{geng2020pqr,amin2016resolving,jacq2019learnfromlearner}, e.g. assuming the true reward is only related to the state~\citep{fu2017airl}. Besides, \cite{peng2019vdb} also shows that, by incorporating regularization like mutual information constraints into reward learning, we can largely improve the generalizability of the reward functions. In this paper, we also introduce regularization to enhance the generalizability of the dynamics reward. 

Despite previous efforts, the performance of offline MBRL algorithms is intrinsically limited by the prediction accuracy of the learned dynamics model itself. In this work, with the aim of obtaining more accurate predictions of dynamics, we employ a strongly generalizable reward of dynamics model, which is learned by IRL, to correct the transitions during the policy learning phase in an offline MBRL algorithm.

\section{Additional Results for Dynamics Reward Learning}

\subsection{Dynamics Reward Learning Algorithm}
\label{appendix:model_reward_learning_algorithm}
Here we present the completed algorithm for dynamics reward learning. Let $\gW = \{D \in \reals^{|\gS|^2 |\gA|}: 0< D (s, a, s^\prime) < 1, \forall (s, a, s^\prime) \in \gS \times \gA \times \gS \}$ and $\gP = \{ P \in \reals^{|\gS|^2 |\gA|}: P(s^\prime|s, a) \geq 0, \forall (s, a, s^\prime) \; ; \sum_{s^\prime \in \gS} P (s^\prime|s, a) = 1, \forall (s, a) \in \gS \times \gA \}$ denote the sets of all feasible discriminators and dynamics models, respectively. Besides, for a set $\gX \subseteq \mathbb{R}^d$, we use $\Pi_{\gX}$ to denote the $\ell_2$-norm-based projection operator onto the set $\gX$, i.e., $\Pi_{\gX} (y) = \argmin_{x \in \gX} \lnorm x - y \rnorm_2^2 $. 
\begin{algorithm}[htbp]
\caption{Dynamics Reward Learning} \label{alg_model_reward_learning} 
\KwIn{Offline data $\mathcal{D}$; number of iterations $T$; step sizes $\{ \eta^{D}_t, \eta^{P}_t \}_{t=1}^T$}
Initialize the discriminator $D_1$ and dynamics model $P_1$\; 
\For {$t= 1, \dots, T$}
{
$D_{t+1} \leftarrow \Pi_{\gW} \bigg( D_t + \eta^{D}_t \big( \expect_{(s, a, s^\prime) \sim \gD} \ls \nabla_{D} \log \lp D_t (s, a, s^\prime) \rp \rs + \expect_{(s, a) \sim \gD, s^\prime \sim P (\cdot|s, a)} \ls \nabla_{D} \log \lp 1-D_t (s, a, s^\prime) \rp \rs  \big) \bigg)$\;\label{alg_line_d_updateing}
$P_{t+1} \leftarrow \Pi_{\gP} \lp P_t - \eta^{P}_t \expect_{(s, a) \sim \gD, s^\prime \sim P (\cdot|s, a)} \ls \nabla_{P} \log \lp P_t (s^\prime|s, a) \rp \log \lp 1 - D_t (s, a, s^\prime) \rp \rs \rp$ \;\label{alg_line_g_updateing}
}
\KwOut{The reward model: $r(s, a, s^\prime) = - \log (1 - \sum_{t=1}^T \log (1-D_t (s, a, s^\prime)) / T)$}
\end{algorithm}

\subsection{Proof of Proposition \ref{prop:convergence}}
\label{app_proposition_prove}
The proof idea is partly inspired by the proof of Lemma 6 in \citep{xu2023provably}. To prove Proposition \ref{prop:convergence}, we need the following auxiliary Lemma.
\begin{lem}
\label{lem:landscape_analysis}
    Consider the objective $\ell (D, P) = \expect_{(s, a, s^\prime) \sim \gD} \ls \log (D (s, a, s^\prime)) \rs + \expect_{(s, a) \sim \gD, s^\prime \sim P (\cdot|s, a)} \ls \log (1-D(s, a, s^\prime)) \rs$. $\ell (D, P)$ is concave in $D \in \gW$ and convex in $P \in \gP$, where $\gW = \{D \in \reals^{|\gS|^2 |\gA|}: 0< D (s, a, s^\prime) < 1 , \forall (s, a, s^\prime) \in \gS \times \gA \times \gS \}$ and $\gP = \{ P \in \reals^{|\gS|^2 |\gA|}: P(s^\prime|s, a) \geq 0, \forall (s, a, s^\prime) \; ; \sum_{s^\prime \in \gS} P (s^\prime|s, a) = 1, \forall (s, a) \in \gS \times \gA \}$. Besides, $\forall D, D^\prime \in \gW, \lnorm D-D^\prime \rnorm_2 \leq 2\sqrt{|\gS|^2 |\gA|}$, $\forall P, P^\prime \in \gP, \lnorm P - P^\prime \rnorm_2 \leq 2 \sqrt{|\gS| |\gA|}$.  
\end{lem}

\begin{proof} [Proof of Lemma \ref{lem:landscape_analysis}]
    First, we apply the second-order condition \citep{boyd2004convex} to verify the concavity of $\ell (D, P)$ with respect to $D$. In particular, we calculate the gradient $\nabla_{D} \ell (D, P) \in \reals^{|\gS|^2 |\gA|}$ with the element
    \begin{align*}
        \nabla_{D (s, a, s^\prime)} \ell (D, P) = \widehat{d} (s, a, s^\prime) \frac{1}{D(s, a, s^\prime)} - \widehat{d} (s, a) P (s^\prime|s, a) \frac{1}{1-D(s, a, s^\prime)}.
    \end{align*}
    With a slight abuse of notations, we use $\widehat{d} (s, a, s^\prime)$ and $\widehat{d} (s, a)$ to denote the empirical distributions from the offline dataset $\gD$. Concretely,
    \begin{align*}
        \widehat{d} (s, a, s^\prime) = \frac{\sum_{(s_t, a_t, s_{t+1}) \in \gD} \indict \{s_t=s, a_t=a, s_{t+1} = s^\prime \}}{|\gD|}, \widehat{d} (s, a) = \frac{\sum_{(s_t, a_t) \in \gD} \indict \{s_t=s, a_t=a \}}{|\gD|}.
    \end{align*}
    We further calculate the Hessian matrix $\nabla^2_{D} \ell (D, P) \in \reals^{|\gS|^2 |\gA| \times |\gS|^2 |\gA|}$. We note that $\nabla^2_{D} \ell (D, P)$ is a diagonal matrix with the principal diagonal elements of
    \begin{align*}
        \nabla^2_{D (s, a, s^\prime)} \ell (D, P) = - \widehat{d} (s, a, s^\prime) \frac{1}{D(s, a, s^\prime)^2} - \widehat{d} (s, a) P (s^\prime|s, a) \frac{1}{(1-D(s, a, s^\prime))^2} < 0.  
    \end{align*}
    Therefore, the Hessian matrix $\nabla^2_{D} \ell (D, P) \in \reals^{|\gS|^2 |\gA| \times |\gS|^2 |\gA|}$ is a negative definite matrix. Based on the second-order condition \citep{boyd2004convex}, we have that $\ell (D, P)$ is concave in $D$. Besides, for any fixed $D \in \gW$, $\ell (D, P)$ is a linear function with respect to $P$. Therefore, $\ell (D, P)$ is convex in $P$.

    Furthermore, $\forall D, D^\prime \in \gW$, we have that $\lnorm D - D^\prime \rnorm_2 \leq 2 \lnorm D \rnorm_2 \leq 2 \sqrt{|\gS|^2 |\gA|}$. Similarly, we have that $\forall P, P^\prime \in \gW$, $\lnorm P - P^\prime \rnorm_2 \leq 2 \lnorm P \rnorm_2 = 2\sqrt{ \sum_{(s, a) \in \gS \times \gA} \sum_{s^\prime} P (s^\prime|s, a)^2 } \leq 2 \sqrt{ \sum_{(s, a) \in \gS \times \gA} (\sum_{s^\prime} P (s^\prime|s, a) )^2 } = 2\sqrt{|\gS| |\gA|}$. We finish the proof.
\end{proof}
Now we proceed to prove Proposition \ref{prop:convergence}. We have that
\begin{align*}
    f \lp \frac{1}{T} \sum_{t=1}^T D_t \rp = \min_{P \in \gP} \ell \lp \frac{1}{T} \sum_{t=1}^T D_t, P  \rp \geq \min_{P \in \gP} \frac{1}{T} \sum_{t=1}^T \ell (D_t, P).  
\end{align*}
The last inequality follows that $\ell (D, P)$ is concave in $D$ from Lemma \ref{lem:landscape_analysis}. Notice that the variable $P$ takes the projected gradient descent updates with respect to the sequence of functions $\{ \ell (D_t, P) \}_{t=1}^T$. From Lemma \ref{lem:landscape_analysis}, we have that $\forall t \in [T]$, $\ell (D_t, P)$ is convex in $P$ and $\forall P, P^\prime \in \gP, \lnorm P - P^\prime \rnorm_2 \leq 2 \sqrt{|\gS| |\gA|}$. Furthermore, we suppose that the gradient norm is bounded, i.e., $\lnorm \nabla_{P} \ell (D_t, P_t) \rnorm_2 \leq G^{P}$. We can apply Theorem 3.1 in \citep{Hazan16introduction-to-oco} to obtain
\begin{align*}
    \min_{P \in \gP} \frac{1}{T} \sum_{t=1}^T \ell (D_t, P) \geq \frac{1}{T} \sum_{t=1}^T \ell (D_t, P_t)   - 3 G^{P} \sqrt{\frac{|\gS| |\gA|}{T}}.
\end{align*}
Then we arrive at
\begin{align*}
    f \lp \frac{1}{T} \sum_{t=1}^T D_t \rp \geq \frac{1}{T} \sum_{t=1}^T \ell (D_t, P_t)   - 3 G^{P} \sqrt{\frac{|\gS| |\gA|}{T}}. 
\end{align*}
Similarly, the variable $D$ takes the projected gradient ascent updates with respect to the sequence of functions $\{ \ell (D, P_t) \}_{t=1}^T$. From Lemma \ref{lem:landscape_analysis}, we obtain that $\forall t \in [T]$, $\ell (D, P_t)$ is concave in $D$ and $\forall D, D^\prime \in \gD$, $\lnorm D - D^\prime \rnorm_2 \leq 2 \sqrt{|\gS|^2 |\gA|}$. Besides, we suppose that the gradient norm is bounded, i.e., $\lnorm \nabla_{D} \ell (D_t, P_t) \rnorm_2 \leq G^{D}$. Applying Theorem 3.1 in \citep{Hazan16introduction-to-oco} yields that
\begin{align*}
    \frac{1}{T} \sum_{t=1}^T \ell (D_t, P_t) \geq \max_{D \in \gW} \frac{1}{T} \sum_{t=1}^T \ell (D, P_t) -  3 G^{D} \sqrt{\frac{|\gS|^2 |\gA|}{T}}. 
\end{align*}
We combine the above two inequalities and get that
\begin{align*}
    f \lp \frac{1}{T} \sum_{t=1}^T D_t \rp &\geq \max_{D \in \gW} \frac{1}{T} \sum_{t=1}^T \ell (D, P_t) - 3 G^{P} \sqrt{\frac{|\gS| |\gA|}{T}} -  3 G^{D} \sqrt{\frac{|\gS|^2 |\gA|}{T}}
    \\
    &\geq \max_{D \in \gW} \min_{P \in \gP} \ell (D, P) - 3 G^{P} \sqrt{\frac{|\gS| |\gA|}{T}} -  3 G^{D} \sqrt{\frac{|\gS|^2 |\gA|}{T}}
    \\
    &= \max_{D \in \gW} f (D) - 3 G^{P} \sqrt{\frac{|\gS| |\gA|}{T}} -  3 G^{D} \sqrt{\frac{|\gS|^2 |\gA|}{T}}. 
\end{align*}
We complete the proof.

\section{Implementation Details}
\subsection{Detailed Implementation of Reward Learning}
\label{app_reward_learning}

We adopt the dynamics reward learning approach grounded on the GAIL framework~\citep{ho2016gail}. As delineated in Section~\ref{sec_reward_learning} and Algorithm~\ref{alg_model_reward_learning}, this process involves two main iterative stages: the discriminator updating stage and the dynamics model updating stage.

During the discriminator updating stage (refer to line~\ref{alg_line_d_updateing} in Algorithm~\ref{alg_model_reward_learning}), the parameters of the discriminator are updated using a single gradient step. This step utilizes both the offline dataset and data procured from the current dynamics model. Conversely, in the dynamics model updating phase, data is sampled from the current model to compute rewards based on $\log(1-D_t(s,a,s'))$. The dynamics model parameters are subsequently refined through an advanced policy gradient method, namely PPO~\citep{schulman2017ppo}. To ensure stable learning dynamics between the discriminator and the dynamics model, the updating of the dynamics model and the discriminator model is repeated $5$ times ($\texttt{d\_step} = \texttt{g\_step} = 5$).  

Additionally, beyond the standard two-term discriminator loss as depicted in \eqref{eq_original_discriminator_loss}, we integrate a gradient penalty to regularize the discriminator, thus curtailing overfitting on offline data. The final discriminator's optimization objective is represented as:
\begin{equation}
    \label{eq_final_discriminator_loss}
    \tilde{\ell}(D, P) =  \ell(D,P) + \eta\Vert\nabla_{D}\ell(D,P)\Vert_2^2,
\end{equation}
where $\eta$ serves as a regularization coefficient.

To further optimize computational efficiency, we cap the stored discriminators to a maximum count of $L$ and archive the discriminator at every $H$ iterations. For tasks in D4RL and NeoRL, the parameters are set at $L=40$ and $H=10$. However, for the refrigerator task, we use $L=200$ and $H=1$. Other hyper-parameters used for dynamics reward learning are listed in Table~\ref{tab_dynamics_reward_learning_parameters}.

\begin{table}[ht]
    \centering
    \small
    \caption{The hyper-parameters of the dynamics reward learning.}
    \begin{tabular}{l|c}\toprule
    \textbf{Attribute} & \textbf{Value} \\
    \midrule
    Number of training iterations & 5000 \\
    Batch size per PPO epoch  & 20000 \\
    Discriminator learning rate & 1e-3 \\
    Adversarial dynamics model learning rate & 3e-4 \\
    Value network learning rate & 1e-3 \\
    Optimizer & Adam~\citep{kingma2015adam} \\ 
    Hidden layers of the adversarial dynamics model & [256, 256] \\
    Hidden layers of the value function & [256, 256] \\
    Hidden layers of the discriminator & [128, 256, 256, 128] \\ 
    $\eta$ & 0.75 for \texttt{HalfCheetah} in NeoRL and 0.5 otherwise  \\
    \bottomrule
    \end{tabular}
    \label{tab_dynamics_reward_learning_parameters}
\end{table}

\subsection{Detailed Implementation of the Uncertainty Estimation in \ours-MOPO}
\label{app_uncertain_estimation}

The proposed \ours-MOPO technique commences by training an ensemble of dynamics models, denoted as $\mathcal{P}=\{p_{\theta_i} (s'| s,a) \mid i\in[M]\}$. Each individual model within this ensemble, specifically $p_{\theta_i}(s'|s,a)$, is characterized as a Gaussian distribution, which is parameterized by a neural network $g(s,a;\theta_i)$. The output of this neural network is split into two components: the mean $g^\mu(s,a;\theta_i)$ and the standard deviation $g^\sigma(s,a;\theta_i)$. Thus, the dynamics model can be explicitly expressed as:
\[ p_{\theta_i}(s'|s,a) = \mathcal{N}(\cdot|g^\mu(s,a;\theta_i), g^\sigma(s,a;\theta_i)^2). \]

To estimate the uncertainty, we employ the concept of max aleatoric~\citep{yu2020mopo}. This term depends on the maximum aleatoric error. In a mathematical context, the aleatoric uncertainty for each model is defined as the L2-norm of its standard deviation, represented as $\Vert g^\sigma(s,a;\theta_i) \Vert_2$ for the $i$-th model. Therefore, the maximal aleatoric error is formalized in \eqref{eq_mopo_uncertainty}:
\begin{equation}
    \label{eq_mopo_uncertainty}
    \mathcal{U}(\mathcal{P}, s_t,a_t,s_{t+1}) = \max_i \Vert g^\sigma(s,a;\theta_i) \Vert_2.
\end{equation}

\subsection{Detailed Implementation of \ours}
We developed \ours using the \texttt{OfflineRL-Kit} codebase\footnote{\url{https://github.com/yihaosun1124/OfflineRL-Kit}}. From this base, we have made the following primary modifications:
\begin{enumerate}
    \item Initialize the program by loading the pre-trained dynamics reward function.
    \item Refine the \texttt{step} method within the dynamics model class:
    \begin{enumerate}
        \item Execute sampling for each model within the ensemble $N$ times.
        \item Compute the dynamics rewards for the obtained samples.
        \item Formulate a softmax distribution and sample an index from this distribution.
        \item If the dynamics reward of the sampled next state is less than $r_{\min}^D$, set the \texttt{terminal} signal to $True$.
        \item Return the next state and the \texttt{terminal} signal corresponding to the sampled index.
    \end{enumerate}
\end{enumerate}

\section{Experiment Details}
\label{app_experiment_details}
\subsection{Detailed Experiment Setting}
\label{app_experiment_setting}
\noindent{\textbf{The refrigerator temperature-control task.}}  In the refrigerator temperature-control task, an agent controls the compressor's power in a refrigerator with the primary aim of maintaining a set target temperature. The agent's observations is the present temperature, with the action corresponding to the normalized power of the compressor.

The system employs a transition function $f(\text{temp}, a, z):\mathbb{R}\times\mathbb{R}\times\{0,1\}\rightarrow\mathbb{R}$ that operates in two distinct modes. The modes indicate the current state of the refrigerator door: open ($z=1$) or closed ($z=0$). These modes are characterized by different rates of temperature change, with the open-door state causing a more rapid convergence towards room temperature (15 $^\circ$C). The transition function, as detailed in \eqref{eq_refrigerator_transition}, formalizes this behavior:
\begin{equation}
\label{eq_refrigerator_transition}
f(\text{temp}_t, a_t, z_t) = \text{temp}_t - a + (0.02 + 0.06z)(15-\text{temp}_t).
\end{equation}
\cref{fig_refrigrator_result} (a) graphically illustrates the dynamics of this system when the refrigerator door is closed ($z=0$), mapping temperature, action, and the next temperature on the x, y, and color axes respectively.

We generated offline data using a proportional controller as the behavior policy, simulating 2000 interactions in an environment where the refrigerator door periodically opens and closes. This policy, formalized in \eqref{eq_refrigerator_sample_policy}, is also depicted in \cref{fig_refrigrator_result} (a):
\begin{equation}
\label{eq_refrigerator_sample_policy}
\pi^\text{sample}(a_t\mid \text{temp}_t)=\mathcal{N}(a_t\mid 0.2\text{temp}_t+0.75, 0.1).
\end{equation}

In the policy learning stage (\cref{fig_refrigerator_policy_vis}, \ref{fig_refrigerator_policy_vis_mu}, and Table~\ref{tab_refrigerator_return}, \ref{table_refri_learned_policy_performance_mu}), we designed a temperature-control task. The corresponding reward is set to 
\begin{equation}
    \label{eq_refrigerator_reward}
    r(\text{temperature}_t) = -|\text{temperature}_t + 1|,
\end{equation}
where $\text{temperature}_t$ is the temperature of the $t$-th step.

\noindent{\textbf{The NeoRL tasks.}} We installed NeoRL from the official repository\footnote{\url{https://github.com/polixir/NeoRL}} and used its $1000$-trajectory offline dataset to accomplish all NeoRL experiments.  
\subsection{Hyper-parameters}

The hyper-parameters for \ours-MOPO and \ours-MOBILE derive from the default parameters specified in MOPO\footnote{\url{https://github.com/yihaosun1124/OfflineRL-Kit}} and MOBILE in \texttt{OfflineRL-Kit}\footnote{\url{https://github.com/yihaosun1124/mobile}}. Modifying these defaults, we extend the rollout horizon from either $1$ or $5$ to $100$, while diminishing the number of rollouts per epoch to alleviate computational strain. This adaptation leads to the consolidated hyper-parameters for both \ours-MOPO and \ours-MOBILE, as detailed in Table~\ref{tab_mopo_hyper}.

Observing that the model transition MAE consistently exceeds $1.0$ when $r_{\min}^D\leq0.6$ as indicated in \cref{fig_d4rl_reward_vis_a}, we establish $r_{\min}^D$ at $0.6$. Notably, this $1.0$ surpasses all validation root mean square errors presented in Table~\ref{tab_d4rl_model_holdout_loss}. Additionally, a value of $0.6$ demonstrates efficacy across a majority of tasks when applied to \ours-MOPO and \ours-MOBILE.

Our investigation primarily revolved around tuning two critical hyper-parameters: the temperature coefficient $\kappa$ and the penalty coefficient $\beta$. 

For $\kappa$, the search space is limited to $\{0, 0.1\}$, where $\kappa=0$ indicates the selection of the subsequent state relying on the maximum dynamics reward.

Regarding $\beta$, we began by identifying the optimal value for \ours-MOPO within the D4RL tasks. We determined the search space for $\beta$ by adjusting the default penalty coefficient using scaling factors $\{0.1, 0.5, 1.0, 2.0, 3.0, 4.0\}$. In the context of NeoRL, given the absence of a predefined $\beta$ in the \texttt{OfflineRL-Kit} repository, we need to search the hyper-parameters in the NeoRL tasks from scratch. Setting the base value of $\beta$ at $2.0$, we then make a search around the base value. It was observed that both \texttt{Hopper-v3-H} and \texttt{Hopper-v3-M} demand a larger $\beta$, whereas \texttt{Walker2d-v3-H} requires a markedly smaller value. The finalized hyper-parameter configurations for \ours-MOPO can be found in Table~\ref{tab_mopo_tasks_hyper}.

Transitioning to \ours-MOBILE, we employed a searching strategy akin to that used in \ours-MOPO. Using the $\beta$ values reported by \cite{sum2023mobile} as a foundation, we conducted multiple searches around this baseline by adjusting $\beta$ using coefficients such as $[0.3, 0.4, 0.5, 0.75, 1.0, 1.5]$. It was discerned that \ours-MOBILE exhibits slightly higher sensitivity to variations in $\beta$ compared to \ours-MOPO. This prompted an additional round of hyper-parameter refinement for some tasks, based on initial search outcomes. The definitive hyper-parameter settings for \ours-MOBILE are detailed in Table~\ref{tab_mobile_tasks_hyper}.

\begin{table}[ht]
    \centering
    \small
    \caption{The common hyper-parameters in \textit{\ours-MOPO} and \textit{\ours-MOBILE}}
    \begin{tabular}{l|c}\toprule
    \textbf{Attribute} & \textbf{Value} \\
    \midrule
    Actor learning rate & 1e-4 \\
    Critic learning rate & 3e-4 \\
    Dynamics learning rate & 1e-3 \\
    Model ensemble size & 7 \\
    Number of the selected models (M in \eqref{eq_ours_tc_generating})  & 5 \\
    Number of resample times (N in \eqref{eq_ours_tc_generating}) & 2 \\
    The number of critics & 2 \\
    Hidden layers of the actor network & [256, 256] \\
    Hidden layers of the critic network & [256, 256] \\
    Hidden layers of the dynamics model  & [200, 200, 200, 200] \\
    Discount factor $\gamma$ & 0.99 \\
    Target network smoothing coefficient $\tau$ & 5e-3 \\
    Max Rollout horizon $H_{\text{rollout}}$ & 100 \\
    Optimizer of the actor and critics & Adam \\
    Rollout number per epoch & 2000 \\
    Batch size of optimization & 256 \\
    Batch number of inferring reward & 4 \\
    Threshold of dynamics reward for rollout termination ($r_{\min}^D$) & 0.6 \\
    Total steps of optimization & 3e6 \\
    Actor optimizer learning schedule & Cosine learning schedule \\
    \bottomrule
    \end{tabular}
    \label{tab_mopo_hyper}
\end{table}
\begin{table}[ht]
\centering
\caption{Particular hyper-parameters of different tasks in \textit{\ours-MOPO}}
\label{tab_mopo_tasks_hyper}
\begin{tabular}{l|cccccc}
\toprule
\textbf{TASK}                         & \textbf{Temperature coefficient $\kappa$} & \textbf{Penalty coefficient $\beta$}\\
\midrule
HalfCheetah-v3-H & 0 & 2.0 \\
HalfCheetah-v3-L & 0 & 2.0 \\
HalfCheetah-v3-M & 0 & 2.0 \\
Hopper-v3-H & 0 & 20.0 \\
Hopper-v3-L & 0 & 2.0 \\
Hopper-v3-M & 0 & 20.0 \\
Walker2d-v3-H & 0 & 0.1 \\
Walker2d-v3-L & 0 & 2.0 \\
Walker2d-v3-M & 0 & 2.0 \\
\midrule
halfcheetah-medium-expert-v2 & 0 & 2.5 \\
halfcheetah-medium-replay-v2 & 0.1 & 0.5 \\
halfcheetah-medium-v2 & 0.1 & 0.5 \\
halfcheetah-random-v2 & 0.1 & 0.5 \\
hopper-medium-expert-v2 & 0 & 15.0 \\
hopper-medium-replay-v2 & 0 & 10.0 \\
hopper-medium-v2 & 0.1 & 15.0 \\
hopper-random-v2 & 0 & 10.0 \\
walker2d-medium-expert-v2 & 0 & 1.25 \\
walker2d-medium-replay-v2 & 0 & 0.25 \\
walker2d-medium-v2 & 0.1 & 0.5 \\
walker2d-random-v2 & 0.1 & 1.0 \\
\bottomrule
\end{tabular}
\end{table}

\begin{table}[ht]
\centering
\caption{Particular hyper-parameters of different tasks in \textit{\ours-MOBILE}}
\label{tab_mobile_tasks_hyper}
\begin{tabular}{l|cc}
\toprule
\textbf{TASK}                         & \textbf{Temperature coefficient $\kappa$} & \textbf{Penalty coefficient $\beta$} \\
\midrule
HalfCheetah-v3-H & 0 & 0.8 \\ % 1.5 8/15
HalfCheetah-v3-L & 0 & 0.4 \\ % 0.5 0.8
HalfCheetah-v3-M & 0.1 & 0.5 \\ % 0.5 1.0
Hopper-v3-H & 0.1 & 2.5 \\ % 2.5 1.0 
Hopper-v3-L & 0 & 0.4 \\ % 2.5 25 / 4
Hopper-v3-M & 0 & 2.0 \\ % 1.5 0.75
Walker2d-v3-H & 0 & 0.04 \\ % 0.5 4 / 50
Walker2d-v3-L & 0 & 0.75 \\ % 2.5 0.3
Walker2d-v3-M & 0 & 0.8 \\ % 2.5 0.32
\midrule
halfcheetah-medium-expert-v2 & 0.1 & 1.0 \\ % 1.0 1.0 
halfcheetah-medium-replay-v2 & 0.1 & 0.2 \\ % 0.5 0.4
halfcheetah-medium-v2 & 0.1 & 0.2 \\ % 0.5 0.4
halfcheetah-random-v2 & 0 & 0.25 \\ % 0.5 0.5
hopper-medium-expert-v2 & 0 & 1.5 \\ % 1.5 1.5
hopper-medium-replay-v2 & 0 & 0.6 \\ % 0.1 6.0 
hopper-medium-v2 & 0.1 & 1.5 \\ % 1.5 1
hopper-random-v2 & 0.1 & 7.5 \\ % 5.0 1.5
walker2d-medium-expert-v2 & 0 & 0.2 \\ % 1.5 2/15
walker2d-medium-replay-v2 & 0 & 0.01 \\ % 0.5 1 / 50
walker2d-medium-v2 & 0 & 0.75 \\ % 1.0 0.75 
walker2d-random-v2 & 0 & 1.0 \\ % 2.0 0.5
\bottomrule
\end{tabular}
\end{table}

\subsection{Baselines}
\label{sec_app_baseline}

Here we introduce the baselines used in our experiments, including model-free offline RL and model-based offline RL. 

\textbf{Model-free offline RL.}
\begin{enumerate}[*]
    \item CQL~\citep{kumar2020conservative} adds penalization to Q-values for the samples out of distribution;
    \item TD3+BC~\citep{fujimoto2021minimalist} incorporates a BC regularization term into the policy optimization objective;
    \item EDAC~\citep{an2021uncertainty} proposed to penalize based on the uncertainty degree of the Q-value.
\end{enumerate}
\textbf{Model-based offline RL.}
\begin{enumerate}[*]
    \item COMBO\citep{yu2021combo} which applies CQL in dyna-style enforces Q-values small on OOD samples;
    \item RAMBO\citep{rigter2022rambo} trains the dynamics model adversarially to minimize the value function without loss of accuracy on the transition prediction;
    \item MOPO~\citep{yu2020mopo} learns a pessimistic value function from rewards penalized with the uncertainty of the dynamics model's prediction;
    \item MOBILE~\citep{sum2023mobile} penalizes the rewards with uncertainty quantified by the inconsistency of Bellman estimations under an ensemble of learned dynamics models.
    \item TT~\citep{janner2021offline} applies a Transformer~\citep{vaswani2017transformer} to modeling distributions over trajectories and uses beam search for planning.
\end{enumerate}

\subsection{Computational Infrastructure}
All experiments were conducted on a workstation outfitted with an Intel Xeon Gold 5218R CPU, $4$ NVIDIA RTX 3090 GPUs, and 250GB of RAM, running Ubuntu 20.04.

\section{Additional Experiment Results}
\subsection{\ours-MU and its Validations}
\label{app_morec_mu}
In the refrigerator temperature-control task, we propose a novel strategy that capitalizes on the dynamics reward by integrating it with RL methods. Specifically, we introduce the variant \textit{\ours-MU} (Model Update), where the model's parameters are iteratively updated through RL to optimize the dynamics reward. This process involves the utilization of PPO~\citep{schulman2017ppo} to maximize the cumulative dynamics reward. In the \ours-MU framework, a given policy employs the current model to generate a batch of rollouts. Each transition within these rollouts is then attributed a dynamics reward. Following this, PPO updates the model's parameters to optimize the accumulated dynamics rewards. This procedure is reiterated over multiple epochs to tailor the model to the policy. Notably, we limit the model ensemble size to a single model in \ours-MU to ensure on-policy sampling.

\begin{figure}[ht]
\centering 
\includegraphics[width=0.4\textwidth]{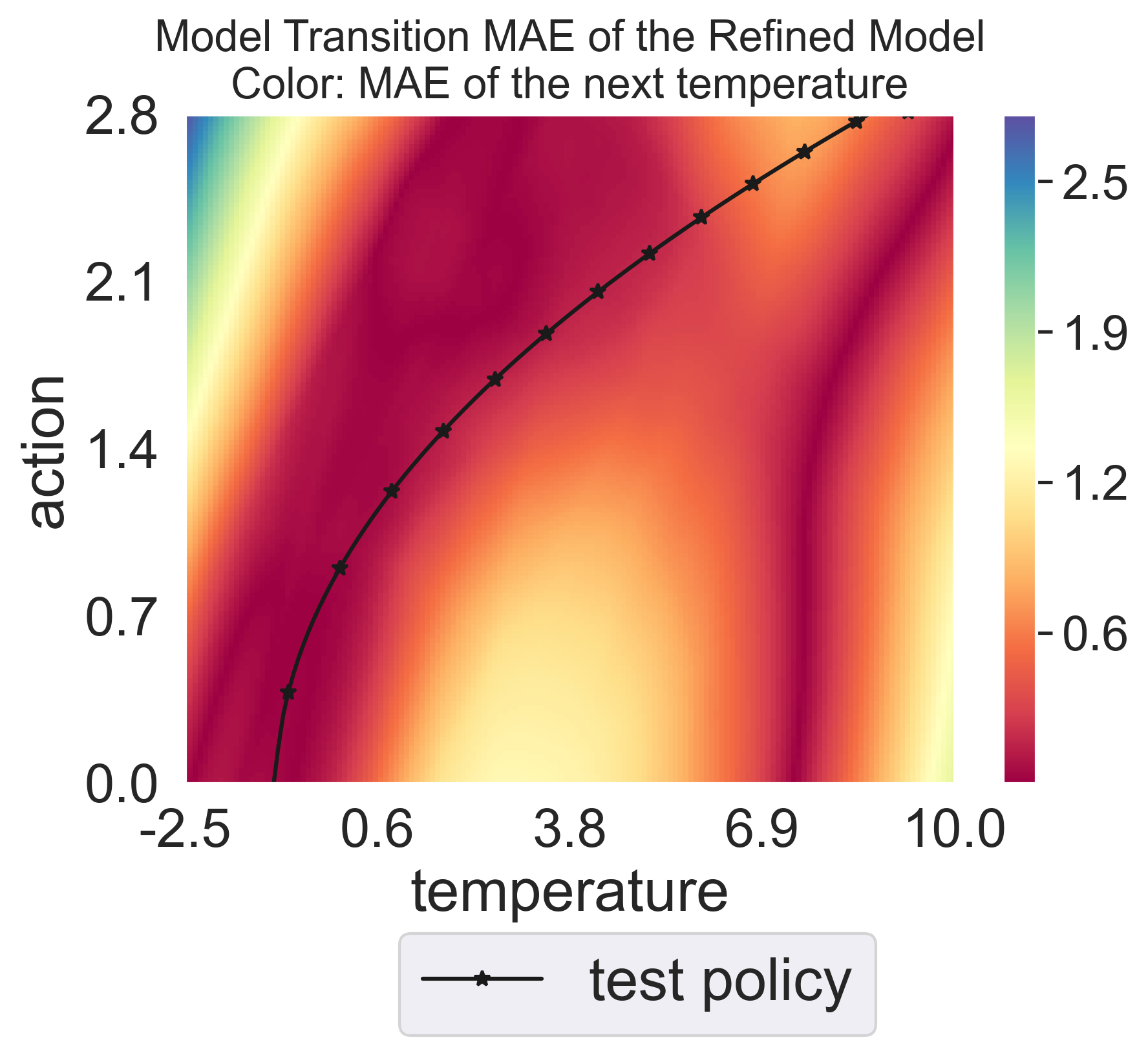}
    \caption{Visualization of the model MAE of the model updated via \ours-MU. The test policy is surrounded by dark red, corresponding to low MAE.} 
    \label{fig_refrigerator_refine_error}
\end{figure}

\begin{figure}[ht]
  \centering
  
  \begin{minipage}{0.45\textwidth}
    \includegraphics[width=\textwidth]{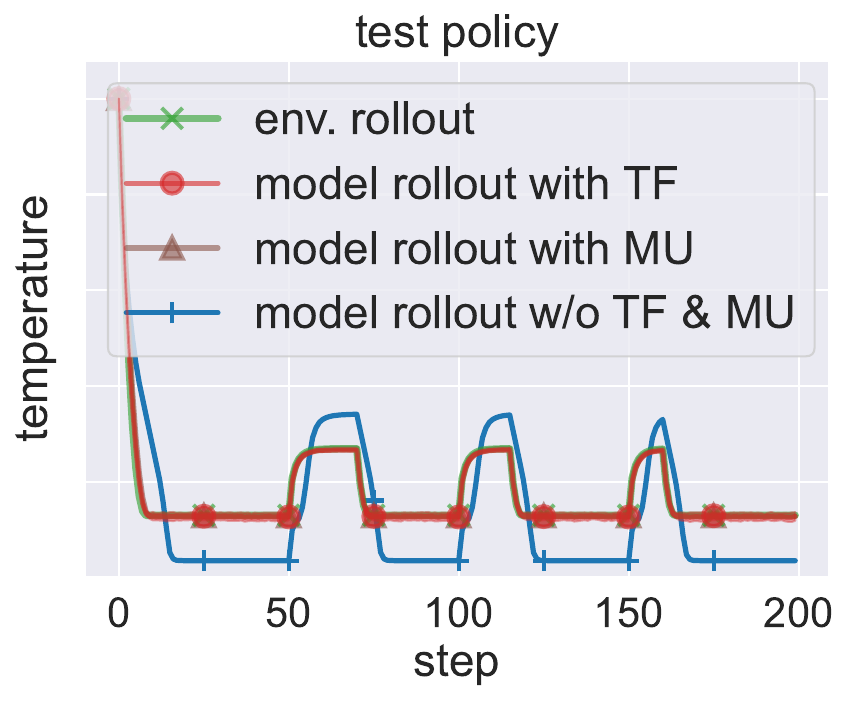}
    \caption{200-step rollouts in model or environment with the test policy.} 
    \label{fig_refrigerator_refine_mu}
  \end{minipage}
  \hspace{1em}
  \begin{minipage}{0.45\textwidth}
    \includegraphics[width=\textwidth]{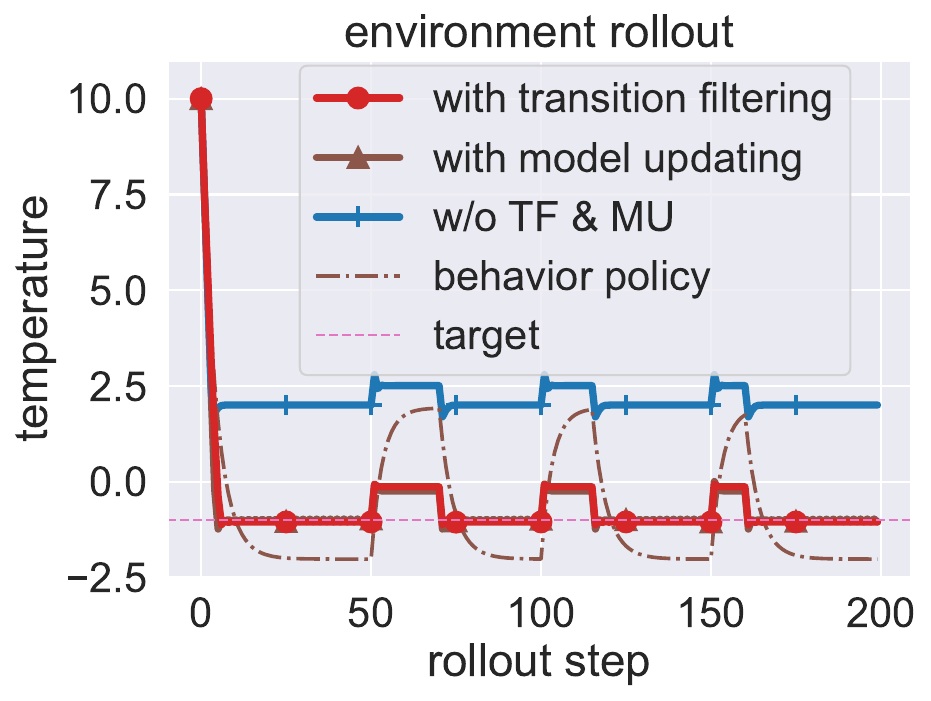}
    \caption{Rollouts of the learned policies in the environment.}
    \label{fig_refrigerator_policy_vis_mu}
  \end{minipage}
\end{figure}

To ascertain the efficacy of the dynamics reward in enhancing model accuracy concerning a test policy, we updated the model, initially trained by \eqref{eq_model_loss}, using \ours-MU, employing 60 PPO updating epochs. The updated model's rollout is depicted in \cref{fig_refrigerator_refine_mu}. It is evident that the rollouts from the \ours-MU updated model closely align with the environment, underscoring its capability to refine the model for unobserved test policies. Further insights are offered in \cref{fig_refrigerator_refine_error}, where the model's MAE is visualized alongside the test policy. Contrasting with \cref{fig_refrigrator_result} (c), the adapted model showcases a significant alteration in the MAE map, with the test policy predominantly situated within the low-MAE zones. This implies that the revised model can offer more precise transitions for the test policy.

\begin{table}[ht]
\centering
\small
  \caption{Mean absolute temperature errors and returns of the learned policies.}
  \label{table_refri_learned_policy_performance_mu}
  \resizebox{1.0\textwidth}{!}{
  \begin{tabular}{lcccc}
    \toprule
     & \textbf{with transition filtering} & \textbf{with model updating} & \textbf{w/o transition filtering \& model updating} & \textbf{behavior policy} \\\midrule
    \textbf{Mean Temperature Error} & $0.39$ & $\mathbf{0.37}$ & $3.20$ & $1.35$ \\
    \textbf{Return} & $-78.0$ & $\mathbf{-74.0}$ & $-640.0$ & $-270.0$ \\
    \bottomrule
  \end{tabular}
  }
  \label{tab_mean_temperature_error_model_update}
\end{table}

Lastly, we seamlessly integrate \ours-MU into the optimal policy learning procedure. We also consider temperature-control task, which aims to stabilize the refrigerator temperature at $1^\circ C$. We adopt an iterative updating strategy comprising two main steps: (i) refining the policy within the current model to maximize the task reward, and (ii) adjusting the model using the present policy to maximize the dynamics reward. This iterative process is performed for a total of $50$ cycles. Specifically, each iteration involves a two-phase update: a $5$-epoch PPO update for the policy, followed by a $5$-epoch PPO update for the dynamics model. The resulting rollouts, generated by the refined policy in the environment, are illustrated in \cref{fig_refrigerator_policy_vis_mu}. Additionally, we provide a quantitative evaluation of the mean temperature error and return in Table~\ref{table_refri_learned_policy_performance_mu}. Evidently, \ours-MU demonstrates superior performance compared to \ours, particularly evident when regulating the temperature during instances like the refrigerator door being open. This is further corroborated by the results presented in Table~\ref{table_refri_learned_policy_performance_mu}, where \ours-MU yields reduced temperature errors.

Collectively, these experimental outcomes highlight an alternative yet promising approach to leveraging the dynamics rewards, showcasing the notable improvement of \ours-MU over \ours. These findings not only advocate for the potential merger of \ours with model updating during policy learning but also emphasize the adaptability of \ours-MU, which can perpetually refine model parameters in alignment with the prevailing policy, thus producing more precise transitions.

\subsection{Memory Cost of the Ensemble Dynamics Model}
\label{supp_memory_cost}

In Algorithm~\ref{alg_model_reward_learning}, we utilize an ensemble approach to formulate the dynamics reward function. Given the necessity to store and manage historical discriminators, one might anticipate a rising memory overhead with \ours. Nonetheless, as demonstrated in Table~\ref{tab_d4rl_memory_cost}, the memory consumption associated with the dynamics reward remains relatively modest. Here, the ensemble size is $40$. 
\begin{table}[ht]
    \centering
    \small
    \caption{Memory cost of the dynamics reward in D4RL tasks.}
    \begin{tabular}{l|r}\toprule
    \textbf{Task Category Name} & \textbf{Memory Cost (MB)} \\\midrule
\texttt{halfcheetah} & 36.843 \\
\texttt{hopper} & 36.257 \\
\texttt{walker2d} & 36.843 \\
\bottomrule
    \end{tabular}
    \label{tab_d4rl_memory_cost}
\end{table}

\subsection{Time Cost Analysis}
To evaluate the computational impact of \ours, we compared the runtime of \ours-MOPO and MOPO with identical hyper-parameters. The mean computational time for each epoch in the \texttt{halfcheetah-medium-v2} task is summarized in Table~\ref{tab_d4rl_time_cost}. Over a span of $3000$ training epochs, \ours-MOPO exhibits an additional overhead of approximately $1.8$ seconds per epoch, translating to an increase of roughly $12.4\%$ in computational time. Consequently, training a \ours-MOPO policy over $3000$ epochs requires an extra $1.5$ hours compared to MOPO. Despite this increase, the notable performance enhancement achieved by \ours-MOPO justifies the additional computational expenditure.

\begin{table}[ht]
    \centering
    \small
    \caption{Time cost of MOPO and MOREC-MOPO in D4RL halfcheetah tasks.}
    \begin{tabular}{lcc}\toprule
    \multirow{2}{*}{\textbf{Task Name}} & \multicolumn{2}{c}{\textbf{Time Cost (s/epoch)}} \\ \cmidrule{2-3}
     & \textbf{MOPO} & \textbf{MOREC-MOPO} \\
    \midrule
\texttt{halfcheetah-medium-v2} & 14.5 & 16.3\\
\bottomrule
    \end{tabular}
    \label{tab_d4rl_time_cost}
\end{table}

\subsection{Performance of the Learned Dynamics Model}
\label{subsec:the_performance_of_the_learned_model}

In accordance with methodologies established in MOPO~\citep{yu2020mopo} and MOBILE~\citep{sum2023mobile}, we also train a suite of dynamic models using supervised learning. This is accomplished by optimizing the log-likelihood, as depicted in \eqref{eq_model_loss}. Importantly, prior to training, we partition our dataset into training and validation subsets. The model exclusively utilizes the training subset for parameter updates. For reference, the validation root mean square errors (RMSEs) are presented in Table~\ref{tab_d4rl_model_holdout_loss}.

\begin{table}[ht]
    \centering
    \small
    \caption{Root mean square error (RMSE) $\pm$ standard error of the model on the validation data, averaged over $5$ seeds.}
    \begin{tabular}{l|r@{~$\pm$~}l}\toprule
& \multicolumn{2}{c}{Model Validation Error}\\\midrule
HalfCheetah-v3-H & $  {0.5273} $ & $  {0.0108} $\\
HalfCheetah-v3-L & $  {0.4164} $ & $  {0.0076} $\\
HalfCheetah-v3-M & $  {0.4917} $ & $  {0.0095} $\\
Hopper-v3-H & $  {0.0615} $ & $  {0.0026} $\\
Hopper-v3-L & $  {0.0878} $ & $  {0.0017} $\\
Hopper-v3-M & $  {0.0664} $ & $  {0.0010} $\\
Walker2d-v3-H & $  {0.4785} $ & $  {0.0069} $\\
Walker2d-v3-L & $  {0.5046} $ & $  {0.0052} $\\
Walker2d-v3-M & $  {0.4488} $ & $  {0.0066} $\\\midrule
halfcheetah-medium-expert-v2 & $  {0.4281} $ & $  {0.0185} $\\
halfcheetah-medium-replay-v2 & $  {0.7182} $ & $  {0.0077} $\\
halfcheetah-medium-v2 & $  {0.4742} $ & $  {0.0071} $\\
halfcheetah-random-v2 & $  {0.3449} $ & $  {0.0039} $\\
hopper-medium-expert-v2 & $  {0.0443} $ & $  {0.0008} $\\
hopper-medium-replay-v2 & $  {0.0734} $ & $  {0.0024} $\\
hopper-medium-v2 & $  {0.0609} $ & $  {0.0027} $\\
hopper-random-v2 & $  {0.0322} $ & $  {0.0041} $\\
walker2d-medium-expert-v2 & $  {0.3361} $ & $  {0.0047} $\\
walker2d-medium-replay-v2 & $  {0.6256} $ & $  {0.0076} $\\
walker2d-medium-v2 & $  {0.5745} $ & $  {0.0070} $\\
walker2d-random-v2 & $  {0.5963} $ & $  {0.0071} $\\
\bottomrule
    \end{tabular}
    \label{tab_d4rl_model_holdout_loss}
\end{table}

\subsection{Additional Performance Results}
\label{app_additional_performance}
We consolidate Table~\ref{tab_d4rl_score} and Table~\ref{tab_neorl_score} into Table~\ref{tab_merge_with_improvement_score}. It is evident from the data that \ours-MOBILE achieves an average score of $80.0$, surpassing the prior SOTA results by a margin of $6.7$. Notably, \ours-MOBILE is the sole method of reaching the score no less than $80.0$. Additionally, it successfully addresses $9$ out of $21$ tasks, marking an $80\%$ enhancement compared to the previous SOTA.

\begin{table}[ht]
\centering
\caption{Performances with maximum improvement ratios in  NeoRL and D4RL benchmarks. Averaged over $5$ seeds. The previous SOTA algorithm is subscripted.}
\label{tab_merge_with_improvement_score}
\resizebox{1.0 \textwidth}{!}{
\begin{tabular}{l|cclr}
\toprule
\textbf{TASK}                         & \textbf{MOREC-MOPO} & \textbf{MOREC-MOBILE} & \textbf{PREV-SOTA} & \textbf{Max Improvement Ratio} \\
\midrule
HalfCheetah-v3-H             & 90.27      & \textbf{98.50}    & 83.00$_{(\text{MOBILE})}$     & 15.50       \\
HalfCheetah-v3-L             & 53.50      & 51.22        & \textbf{54.70}$_{(\text{MOBILE})}$ & -1.20       \\
HalfCheetah-v3-M             & 84.08      & \textbf{85.99}    & 77.80$_{(\text{MOBILE})}$     & 8.19        \\
Hopper-v3-H                  & 72.77      & \textbf{108.48}   & 87.80$_{(\text{MOBILE})}$     & 20.68       \\
Hopper-v3-L                  & 25.36      & \textbf{26.75}    & 18.30$_{(\text{MOBILE})}$     & 8.45        \\
Hopper-v3-M                  & 83.50      & \textbf{109.23}   & 70.30$_{(\text{TD3+BC})}$     & 38.93       \\
Walker2d-v3-H                & \textbf{83.04}  & 79.30        & 74.90$_{(\text{EDAC})}$     & 8.14        \\
Walker2d-v3-L                & \textbf{65.01}  & 56.81        & 44.70$_{(\text{CQL})}$     & 20.31       \\
Walker2d-v3-M                & \textbf{76.59}  & 71.07        & 62.20$_{(\text{MOBILE})}$     & 14.39       \\\midrule
halfcheetah-medium-expert-v2 & \textbf{112.07} & 110.94       & 108.20$_{(\text{MOBILE})}$    & 3.87        \\
halfcheetah-medium-replay-v2 & \textbf{76.45}  & 76.40        & 71.70$_{(\text{MOPO})}$     & 4.75        \\
halfcheetah-medium-v2        & \textbf{82.27}  & 82.12        & 77.90$_{(\text{MOBILE})}$     & 4.37        \\
halfcheetah-random-v2        & 51.57      & \textbf{53.19}    & 39.30$_{(\text{MOBILE})}$     & 13.89       \\
hopper-medium-expert-v2      & \textbf{113.25} & 111.47       & 112.60$_{(\text{MOBILE})}$    & 0.65        \\
hopper-medium-replay-v2      & 105.15     & \textbf{105.45}   & 103.90$_{(\text{MOBILE})}$    & 1.55        \\
hopper-medium-v2             & 106.96     & \textbf{107.97}   & 106.60$_{(\text{MOBILE})}$    & 1.37        \\
hopper-random-v2             & \textbf{32.10}  & 26.58         & 31.90$_{(\text{MOBILE})}$     & 0.20        \\
walker2d-medium-expert-v2    & \textbf{115.77} & 115.52       & 115.20$_{(\text{MOBILE})}$    & 0.57        \\
walker2d-medium-replay-v2    & 95.50      & \textbf{95.83}    & 89.90$_{(\text{MOBILE})}$     & 5.93        \\
walker2d-medium-v2           & 89.93  & 85.76        & \textbf{92.50}$_{(\text{EDAC})}$     & -2.57       \\
walker2d-random-v2           & \textbf{23.53}  & 22.75        & 16.60$_{(\text{MOBILE})}$     & 6.93        \\\midrule
Average &78.0&\textbf{80.0}&73.3&8.3\\
% Average (PREV-SOTA $<$ 100) &67.8&\textbf{68.9}&62.1&10.4\\
Solved tasks (performance $>$ 95) & 6/21 & \textbf{9/21} & 5/21 & -\\
\bottomrule
\end{tabular}
}
\end{table}

\begin{figure}[ht]
\centering 
\includegraphics[width=0.9 \linewidth]{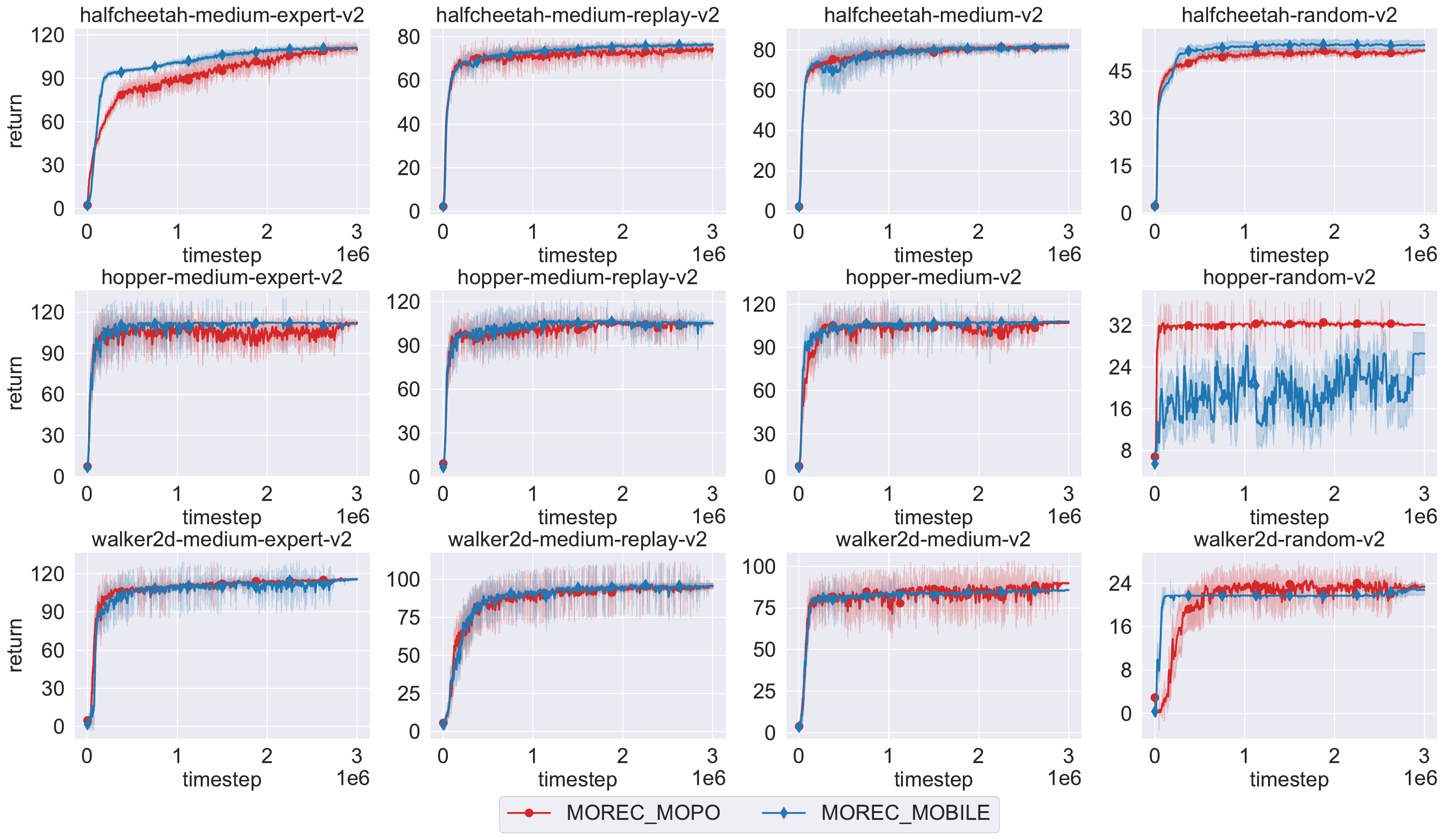}
\caption{The learning curves of \ours-MOPO and \ours-MOBILE in D4RL tasks. The curves are shaded with 1 standard error over $5$ seeds.} 
\label{d4rl_learning_curve}  
\end{figure}

\begin{figure}[ht]
\centering 
\includegraphics[width=0.9 \linewidth]{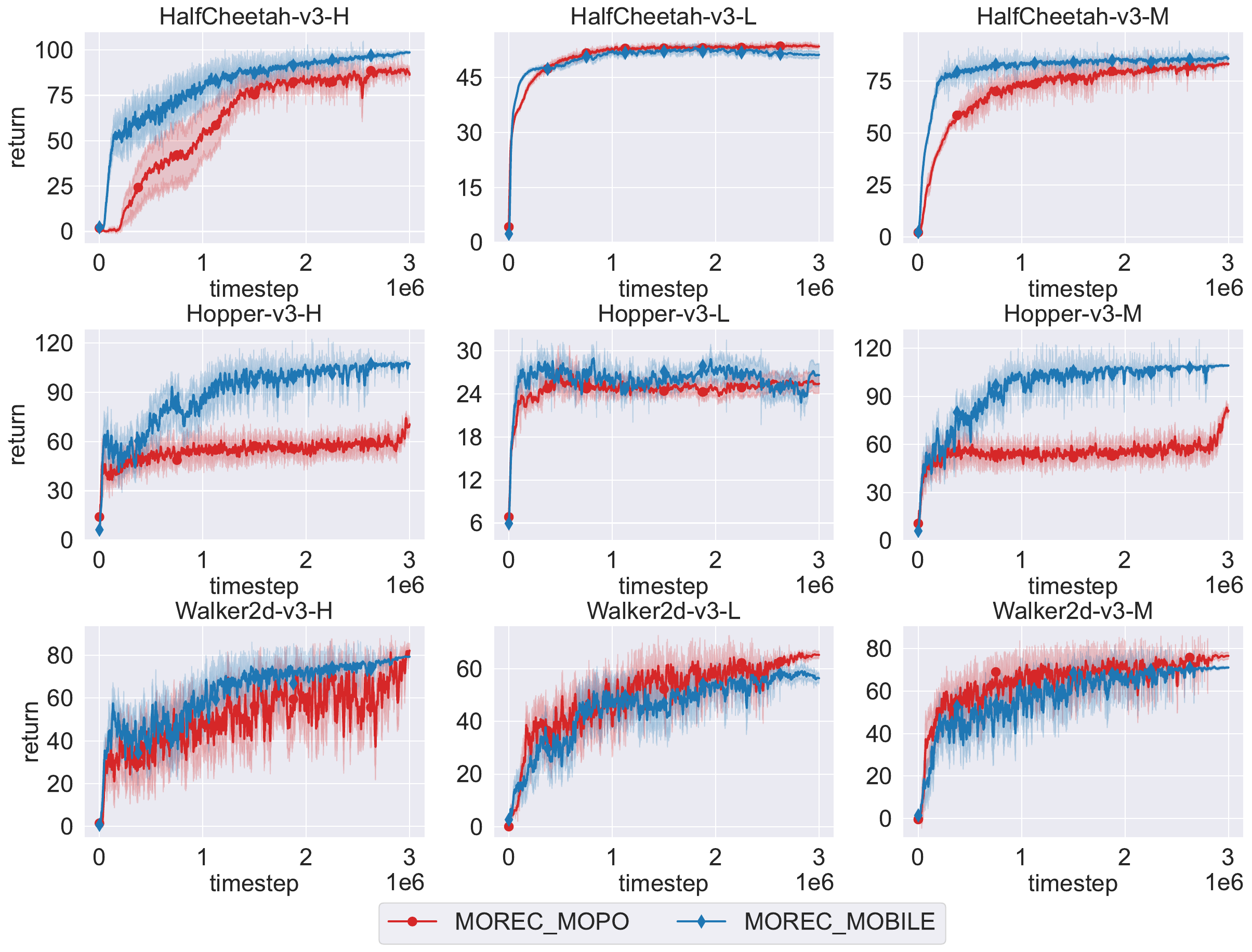}
\caption{The learning curves of \ours-MOPO and \ours-MOBILE in NeoRL tasks. The curves are shaded with one standard error over $5$ seeds.} 
\label{neorl_learning_curve}  
\end{figure}

The learning curves of both \ours-MOPO and \ours-MOBILE are depicted in \cref{d4rl_learning_curve} and \cref{neorl_learning_curve}. These curves indicate that both methodologies maintain a stable learning progression. Furthermore, the small shaded region, representing the standard error, underscores the robustness of \ours against variability in random seed initialization.

\subsection{Additional Ablation Studies}
\label{app_ablation_studies}
To discover the efficacy of the transition filtering technique, we devised a variant of \ours, devoid of this technique, i.e., \textit{w/o transition filtering}. The hyper-parameters for this variant remain consistent with those of \ours. Subsequently, we trained policies using both methods on the \texttt{hfctah} tasks from the D4RL benchmark. Comparative results of the policy performance are detailed in Table~\ref{tab_d4rl_ablation}. A noticeable degradation in the performance of \ours-MOPO without transition filtering underlines the pivotal role of this technique.

\begin{table}[ht]
    \centering
    \small
    \caption{The final normalized policy returns of \textit{\ours-MOPO} and a variant of \textit{\ours-MOPO} without transition filtering in \texttt{hfctah} tasks, averaged over $5$ seeds.}
    \begin{tabular}{l|cccc}\toprule
    & \textbf{rnd} & \textbf{med} & \textbf{med-rep} & \textbf{med-exp} \\
    \midrule
    \textbf{\ours-MOPO} & $\mathbf{51.57}\pm \mathbf{0.5}$ & $\mathbf{82.3}\pm \mathbf{1.1}$ & $\mathbf{76.5}\pm \mathbf{1.2}$ & $\mathbf{112.1}\pm \mathbf{1.8}$ \\
    \textbf{w/o transition filtering} & $38.42\pm 2.1$ & $44.18\pm 4.5$ & $73.3\pm 1.5$ & $80.4\pm 9.2$ \\
    \bottomrule
    \end{tabular}
    \label{tab_d4rl_ablation}
\end{table}
Besides, we additionally consider the maximum transition MAE. For each model rollout, we obtain all its transition MAEs and choose the maximum one for this rollout. Then, we take the average of these maximum MAEs across all rollouts collected during the full training process and obtain the \textit{maximum transition MAE}. We list the maximum transition MAE for 12 D4RL tasks in Table~\ref{tab_d4rl_max_rollout_mae}, showing a significant MAE reduction of \ours-MOPO. This result implies \ours-MOPO can avoid inserting the outlier transitions to its replay buffer and ensure the training stability. 
\begin{table}[ht]
    \centering
    \small
    \caption{The average maximum transition MAE $\pm$ standard error, averaged over $5$ seed, in 12 D4RL tasks. For each model rollout, we obtain all its transition MAEs and choose the maximum one for the model rollout. Then, we take the average of these maximum MAEs across all rollouts and obtain the maximum transition MAE.}
    \begin{tabular}{l|r@{~$\pm$~}lr@{~$\pm$~}l}\toprule
& \multicolumn{2}{c}{\textbf{\ours-MOPO}} & \multicolumn{2}{c}{\textbf{w/o transition filtering}}\\\midrule
halfcheetah-medium-expert-v2 & $ \mathbf{6.066} $ & $ \mathbf{1.777} $ & $249906.304$& $48324.466$\\
halfcheetah-medium-replay-v2 & $ \mathbf{3.860} $ & $ \mathbf{0.102} $ & $11.653$& $5.191$\\
halfcheetah-medium-v2 & $ \mathbf{3.282} $ & $ \mathbf{0.049} $ & $304736.147$& $71894.240$\\
halfcheetah-random-v2 & $ \mathbf{3.137} $ & $ \mathbf{0.252} $ & $212908.396$& $33037.017$\\
hopper-medium-expert-v2 & $ \mathbf{0.556} $ & $ \mathbf{0.024} $ & $1.553$& $0.300$\\
hopper-medium-replay-v2 & $ \mathbf{0.598} $ & $ \mathbf{0.005} $ & $0.613$& $0.014$\\
hopper-medium-v2 & $ \mathbf{0.529} $ & $ \mathbf{0.004} $ & $0.581$& $0.018$\\
hopper-random-v2 & $ \mathbf{0.183} $ & $ \mathbf{0.003} $ & $0.185$& $0.004$\\
walker2d-medium-expert-v2 & $ \mathbf{3.411} $ & $ \mathbf{0.160} $ & $8.453$& $0.660$\\
walker2d-medium-replay-v2 & $ \mathbf{5.619} $ & $ \mathbf{0.130} $ & $8.568$& $0.570$\\
walker2d-medium-v2 & $ \mathbf{3.892} $ & $ \mathbf{0.178} $ & $4.532$& $0.360$\\
walker2d-random-v2 & $ \mathbf{6.964} $ & $ \mathbf{0.601} $ & $12.640$& $1.344$\\
\bottomrule
    \end{tabular}
    \label{tab_d4rl_max_rollout_mae}
\end{table}

\subsection{Additional Dynamics Reward Differences Visualizations}
\label{app_additional_reward_difference}

\begin{figure}[ht]
\centering 
\includegraphics[width=1.0 \linewidth]{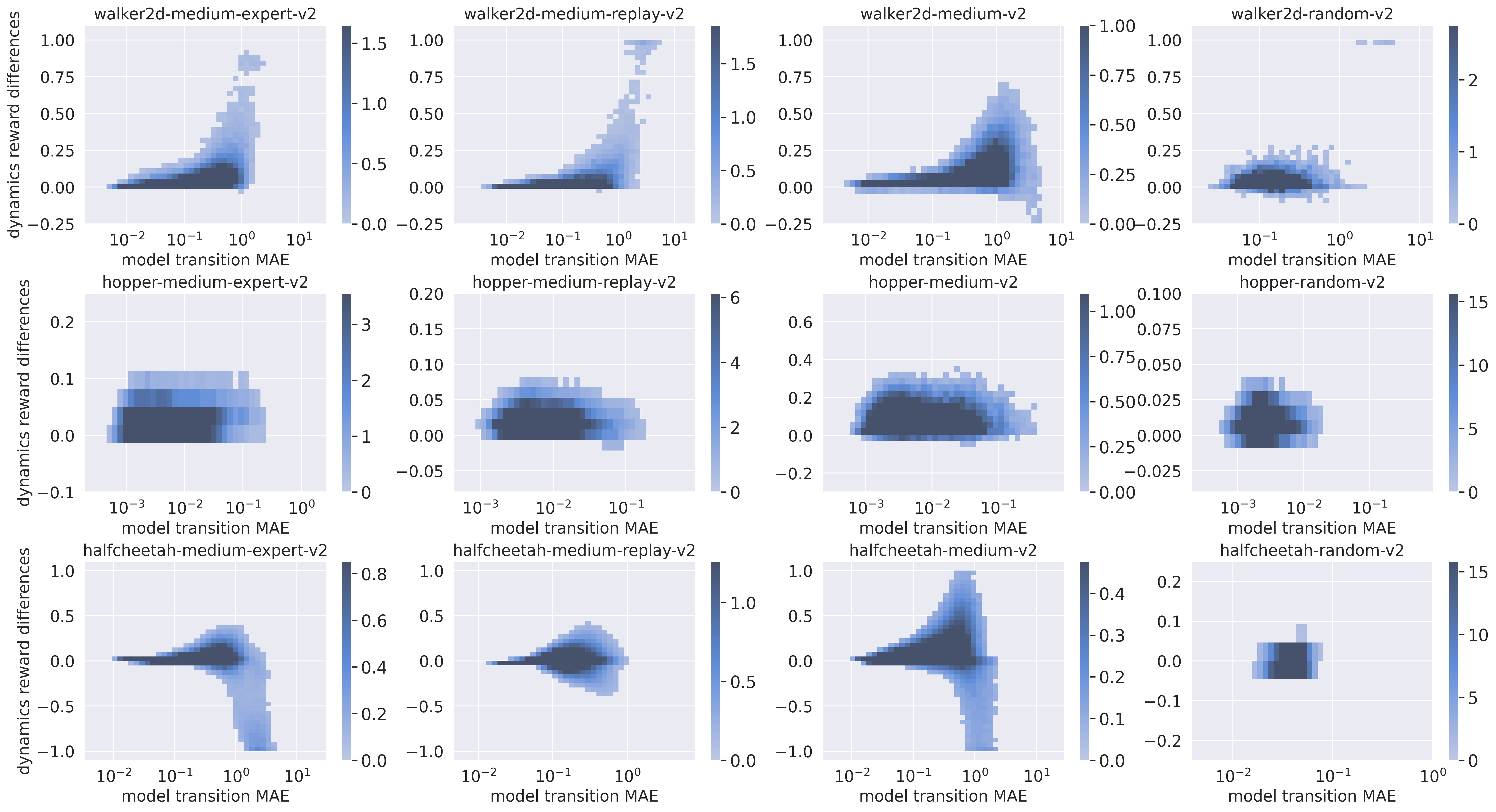}
\caption{The joint distribution of the dynamics reward differences $r^D(s_t, a_t, s_{t+1}^\star)-r^D(s_t,a_t, s_{t+1})$ and the model transition MAEs $\Vert s_{t+1} - s_{t+1}^\star \Vert_1$ in 12 D4RL tasks.} 
\label{reward_diff_total}  
\end{figure}

We analyze the relationship between dynamics rewards and the model transition MAE, denoted as $\Vert s_{t+1} - s_{t+1}^\star \Vert_1$, across $12$ D4RL tasks as shown in \cref{reward_diff_total}. Utilizing policies derived from behavior cloning, we sample trajectories from each learned dynamics model, without transition filtering. The joint distribution of $\{(\Vert s_{t+1} - s_{t+1}^\star\Vert, r^D(s_t,a_t,s_{t+1}^\star)-r^D(s_t,a_t,s_{t+1}))\}$ is depicted in \cref{reward_diff_total}, with $r^D(\cdot)$ representing the dynamics reward, $s_{t+1}^\star$ the true next state, and $s_{t+1}$ the next state predicted by the model. It is noteworthy that trajectories in the \texttt{random} data subset are drawn from a random policy, resulting in a tighter distribution compared to other data sources.

In tasks such as \texttt{walker2d} and \texttt{hopper}, the discrepancies in dynamics rewards are predominantly positive, underscoring the potent discriminative prowess of the dynamics reward. This indicates that, for these tasks, the dynamics reward consistently assigns higher values to true transitions over model transitions. The distribution patterns for \texttt{walker2d} are distinct from those of \texttt{hopper}. This distinction stems from the variation in one-step model errors between these tasks, as elaborated in Table~\ref{tab_d4rl_model_holdout_loss}. The root mean square error (RMSE) for \texttt{walker2d} is nearly an order of magnitude greater than \texttt{hopper}. Consequently, cumulative errors in \texttt{hopper} are more contained, leading to a smaller $x$-axis range. Specifically, the $x$-axis span for $\texttt{hopper}$ is roughly $[0.001,0.5]$, in contrast to $[0.008, 7]$ for $\texttt{walker2d}$. In \texttt{walker2d}, dynamics rewards begin to exhibit significant differences when MAE approaches $0.3$, which aligns with the maximum transition MAE observed in \texttt{hopper}. This masks the potential trend of increasing dynamics reward discrepancies with escalating transition MAEs.

\begin{figure}[ht]
\centering 
\includegraphics[width=1.0 \linewidth]{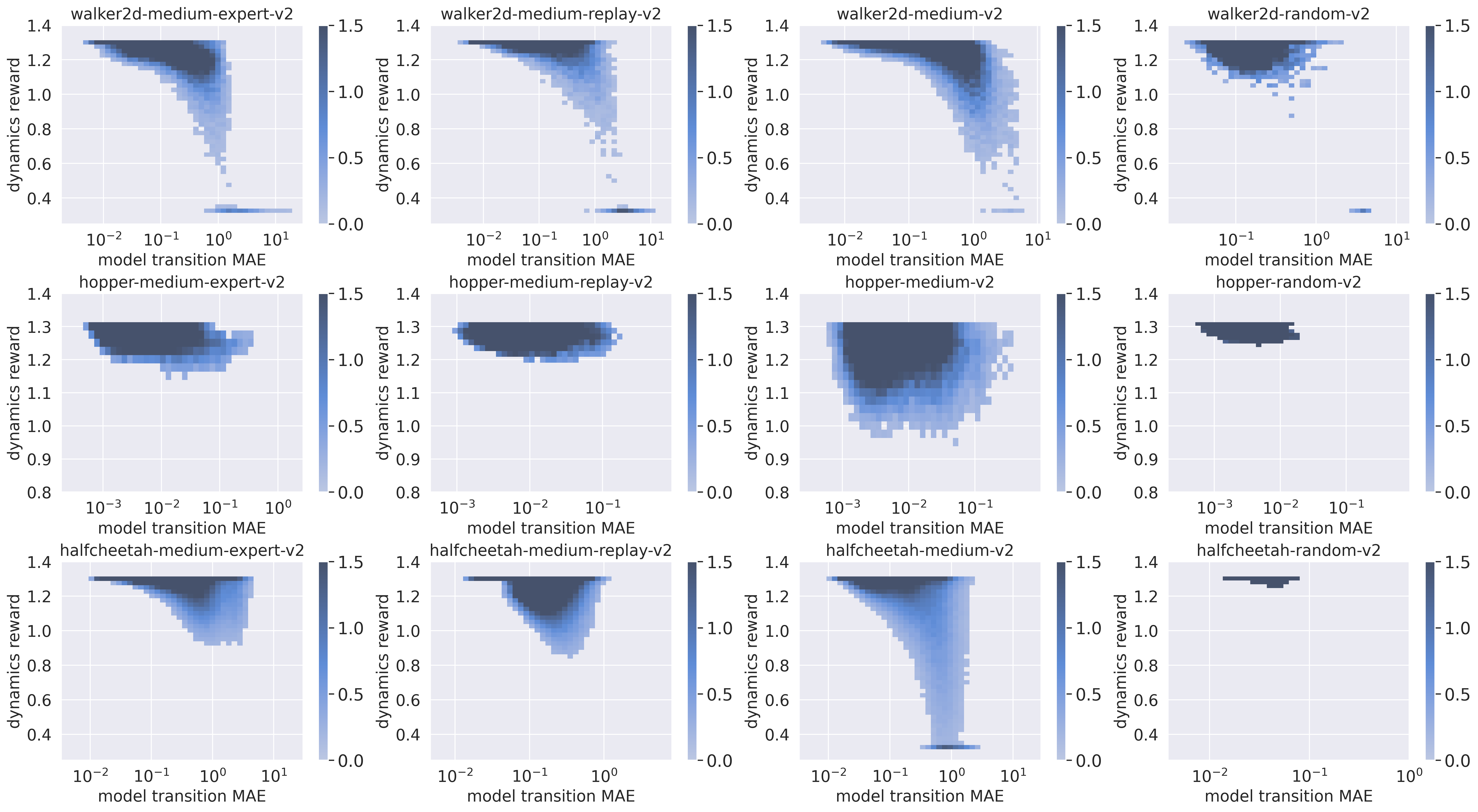}
\caption{The joint distribution of the dynamics reward for model transitions $r^D(s_t,a_t, s_{t+1})$ and the model transition MAEs $\Vert s_{t+1} - s_{t+1}^\star \Vert_1$ in 12 D4RL tasks.} 
\label{model_dynamics_reward_halfcheetah}  
\end{figure}

As for the \texttt{halfcheetah} tasks, on the contrary, the dynamics reward function sometimes gives a higher reward to the model transition than the true transition, which is not desired. In order to discover why \ours also work in the \texttt{halfcheetah} tasks, we additionally visualize the joint distribution of $\{(\Vert s_{t+1} - s_{t+1}^\star\Vert, r^D(s_t,a_t,s_{t+1}))\}$ in \cref{model_dynamics_reward_halfcheetah}. We find the dynamics reward of \texttt{halfcheetah} shows a similar trend as \texttt{walker2d}: As the model transition MAE increases, the dynamics rewards reduce gradually. Because the behavior policy of \texttt{halfcheetah-random-v2} is a random policy, its distribution is narrow and cannot show the aforementioned trend. As a result, the dynamics reward can still give a low dynamics reward to the high-MAE transitions. Consequently, the rollout can be terminated on time before the MAE diverges.  

\end{document}